\documentclass[10pt,twocolumn,letterpaper]{article}

\usepackage{wacv}
\usepackage{times}
\usepackage{epsfig}
\usepackage{graphicx}
\usepackage{amsmath}
\usepackage{amssymb}
\usepackage{booktabs}

\usepackage{caption}
\usepackage{booktabs}
\usepackage{algorithm}
\usepackage{algpseudocode}
\usepackage{varwidth}
\usepackage{array, makecell}
\usepackage[rgb]{xcolor}
\definecolor{OliveGreen}{rgb}{0,0.6,0}
\definecolor{Burgundy}{RGB}{144,0,32}

\newcommand{\authorname}[1]{\fontsize{11}{11}\selectfont #1}
\makeatletter
\def\@fnsymbol#1{\ensuremath{\ifcase#1\or \dagger\or \ddagger\or
   \mathsection\or \mathparagraph\or \|\or **\or \dagger\dagger
   \or \ddagger\ddagger \else\@ctrerr\fi}}
\makeatother

\def\wacvPaperID{1546} %
\def\confName{WACV}
\def\confYear{2024}

\usepackage[pagebackref=true,breaklinks=true,letterpaper=true,colorlinks,linkcolor=.,bookmarks=false]{hyperref}
\usepackage[capitalize]{cleveref}

\begin{document}

\title{Shape-Guided Diffusion with Inside-Outside Attention}

\author{
\bf\authorname{Dong Huk Park$^{1*}$ \\
\bf\authorname{Xihui Liu}$^{1,3}$\thanks{This work was done when Xihui Liu was a postdoc at UC Berkeley.} \\\\
\textsuperscript{1}\authorname{UC Berkeley} \\
\and
\bf\authorname{Grace Luo}$^{1*}$ \\
\bf\authorname{Maka Karalashvili}$^{4}$ \\\\
\textsuperscript{2}\authorname{Meta AI} \\
\and
\bf\authorname{Clayton Toste}$^{1}$ \\
\bf\authorname{Anna Rohrbach}$^{1}$ \\\\
\textsuperscript{3}\authorname{The University of Hong Kong} \\
\and
\bf\authorname{Samaneh Azadi}$^{2}$ \\
\bf\authorname{Trevor Darrell}$^{1}$ \\\\
\textsuperscript{4}\authorname{BMW Group} \\
\and
\small{\textit{$^*$Denotes equal contribution.}}
}
}

\addtolength{\abovecaptionskip}{-0.5cm}
\addtolength{\belowcaptionskip}{-0.5cm}

\makeatletter
\let\@oldmaketitle\@maketitle
\renewcommand{\@maketitle}{\@oldmaketitle
\vspace{-5pt}
\centering
\includegraphics[width=\linewidth]{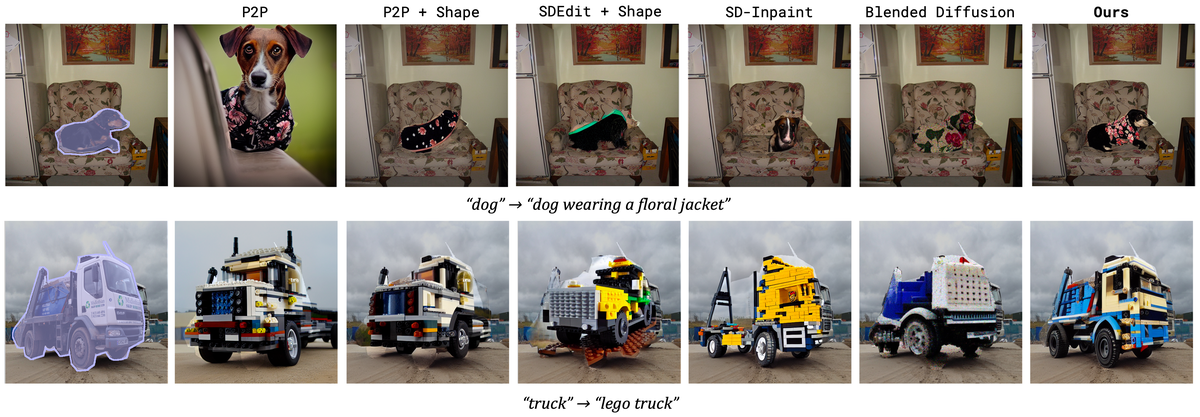}
\vspace{0cm}
\captionof{figure}{We demonstrate that prior work in local image editing \cite{hertz2022prompt, meng2022sdedit, sdinpaint, avrahami2022blended} fails to preserve precise object silhouette. We propose \textit{Shape-Guided Diffusion}, a training-free method that uses a novel \textit{Inside-Outside Attention} to respect shape input.
Our method can be provided an object mask as input or infer a mask from text.}
\label{fig:intro}
\bigskip\bigskip}
\makeatother
\maketitle

\begin{abstract}
   We introduce precise object silhouette as a new constraint in text-to-image diffusion models, which we dub Shape-Guided Diffusion. Our training-free method uses an Inside-Outside Attention mechanism during the inversion and generation process to apply a shape constraint to the cross- and self-attention maps. Our mechanism designates which spatial region is the object (inside) vs. background (outside) then associates edits to the correct region. We demonstrate the efficacy of our method on the shape-guided editing task, where the model must replace an object according to a text prompt and object mask. We curate a new ShapePrompts benchmark derived from MS-COCO and achieve SOTA results in shape faithfulness without a degradation in text alignment or image realism according to both automatic metrics and annotator ratings. Our data and code will be made available at \url{https://shape-guided-diffusion.github.io}.
\end{abstract}

\begin{figure*}
\vspace{-0.3cm}
\begin{scriptsize}
\begin{center}
\includegraphics[width=0.75\linewidth]{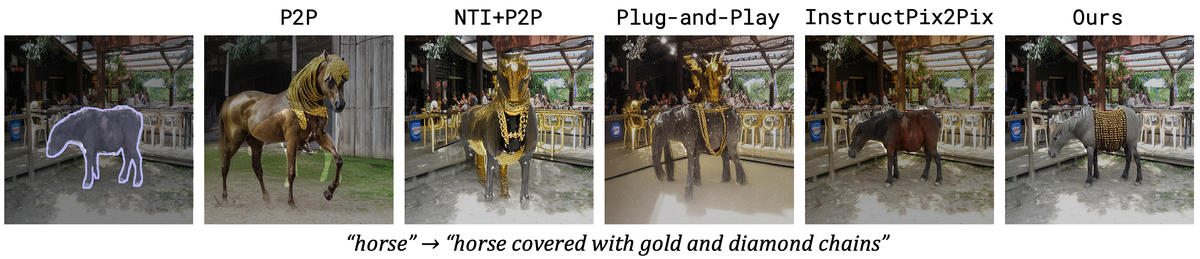}
\end{center}
\captionof{figure}{Our work differs from concurrent work in structure-preserving editing~\cite{mokady2022nulltext, tumanyan2022pnp, brooks2022instructpix2pix} in that we \textit{constrain} attention maps such that edits are \textit{localized} to a spatial region. Here we infer our shape constraint from the text prompt, thereby using the same amount of input as other methods.
}
\label{fig:concurrent-comparison}
\end{scriptsize}
\end{figure*}

\section{\label{sec:intro} Introduction}
\vspace{-0.3cm}
What is \textit{shape}? By definition, an object's shape denotes the boundary, outline, or contour that separates it from the external world. As a result, shapes often carry a great deal of semantic meaning. For example, in the bottom row of ~\autoref{fig:intro} the silhouette alone reveals that the object is a vehicle oriented rightwards,
without the cues of color or texture. Given that object silhouette plays a key role in human visual processing, including object recognition and categorization~\cite{rosch1976basic}, it follows that shape presents as a powerful cue for representing user intent when interacting with generative models. However, prior work in local image editing~\cite{avrahami2022blended, sdinpaint} typically focuses on the coarsest form of shape input, often in the form of amorphous blobs or ``user scribbles'' where it is difficult to discern even the object category from the silhouette alone. As a result, these methods often fail when given precise shape inputs. We instead focus on precise object masks, which are easily acquired from off-the-shelf segmentation models. Thus, we consider the task of \emph{shape-guided editing}, where a real image, text prompt, and object mask are fed to a pre-trained text-to-image diffusion model to synthesize a new object faithful to the the text prompt and the mask's shape.

Our method is motivated by the observation that diffusion models often contain spurious attentions that weakly associate object and background pixels, which makes it difficult to produce an edit that preserves a given shape boundary. To overcome this issue, we delineate the object (inside) and background (outside), with a novel Inside-Outside Attention mechanism that modifies the cross- and self- attention maps such that a token or pixel referring to the object is constrained to attend to pixels inside the shape, and vice versa. We apply this mechanism not only to the generation process to perform shape sensitive edits, but we also apply it to the inversion process to better preserve information about the source object before editing.

To summarize, our contributions include the following:\\
(1) We identify a limitation in prior image editing methods where the shape of the original object is not preserved and provide empirical insights on why this issue exists.\\
(2) Unlike existing mask-based editing adaptations (e.g., copying the background or finetuning the model to use mask input),
we introduce a training-free mechanism that applies a shape constraint on the attention maps at inference time. To the best of our knowledge, we are the first work to explore \textit{constraining} attention maps during inversion, which allows us to discover inverted noise that better preserves shape information from a real image. \\
(3) Our method achieves SOTA results in shape faithfulness on our MS-COCO ShapePrompts benchmark, and is rated by annotators as the best editing method 2.7x more frequently than the most competitive baseline. We demonstrate diverse editing capabilities such as object edits, background edits, and simultaneous inside-outside edits.
\section{\label{sec:related} Related Work}

\noindent\textbf{Diffusion Models}
Diffusion models~\cite{sohl2015deep} 
define a Markov chain of diffusion steps that slowly adds random noise to data then learn a model to reverse this process.
Variants 
include Denoising Diffusion Probabilistic Models (DDPM)~\cite{ho2020denoising}, Denoising Diffusion Implicit Models (DDIM)~\cite{song2020denoising}, and score-based models~\cite{song2020score}.
Recently, diffusion models~\cite{ramesh2021zero,nichol2022glide, rombach2021highresolution,saharia2022imagen} have shown impressive performance on text-guided image synthesis. Our work focuses on adapting these diffusion models towards text-guided local editing according to a text prompt and object mask.

\begin{figure*}
\vspace{-0.2cm}
\begin{scriptsize}
\begin{center}
\includegraphics[width=0.65\linewidth]{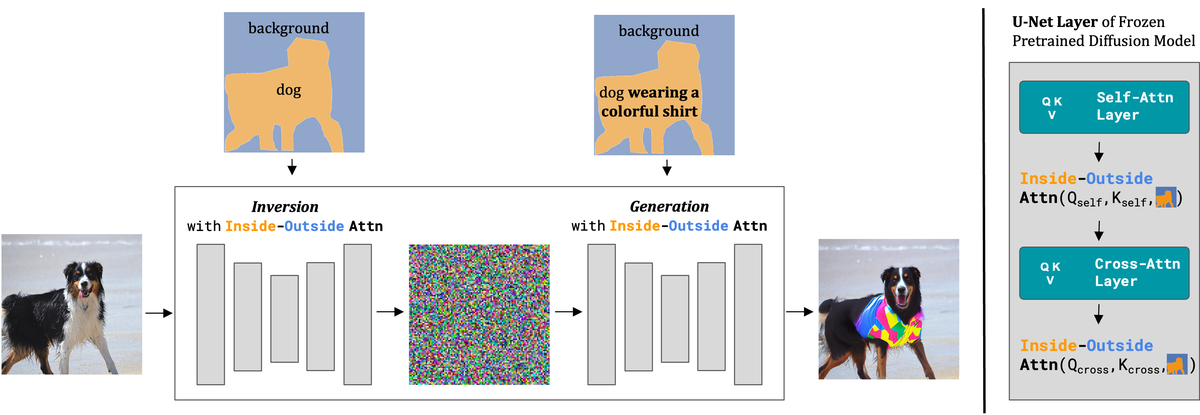}
\end{center}
\caption{Shape-Guided Diffusion. Our method takes a real image, source prompt (``dog''), edit prompt (``dog wearing a colorful shirt''), as well as an optional object mask, and outputs an edited image. We infer the object mask from the source prompt if it is not provided using a shape inference function, e.g., a segmentation model. Left: we modify a frozen pretrained text-to-image diffusion model during both the inversion and generation processes. Right: we show a detailed view of one layer in the U-Net, where Inside-Outside Attention constrains the self- and cross-attention maps according to the mask.}
\label{fig:pipeline}
\end{scriptsize}
\end{figure*}

\noindent\textbf{Global and Local Image Editing}
Various works have extended generative models towards image editing.
For text-guided global editing, StyleCLIP~\cite{patashnik2021styleclip} adapts StyleGAN~\cite{karras2019style} and DiffusionCLIP~\cite{kim2022diffusionclip} adapts diffusion models to edit entire images according to a text prompt. Blended Diffusion \cite{avrahami2022blended} proposes a method for local editing constrained to a mask
by copying an appropriately noised version of the source image's background at each diffusion timestep. 
While this ``copy background'' technique can be generally combined with other methods to enable local editing in diffusion models, we demonstrate that this method alone is insufficient for preserving object shape, and we further improve shape faithfulness with our proposed method.

\begin{figure*}
\vspace{-0.3cm}
\begin{scriptsize}
\begin{center}
\includegraphics[width=0.75\linewidth]{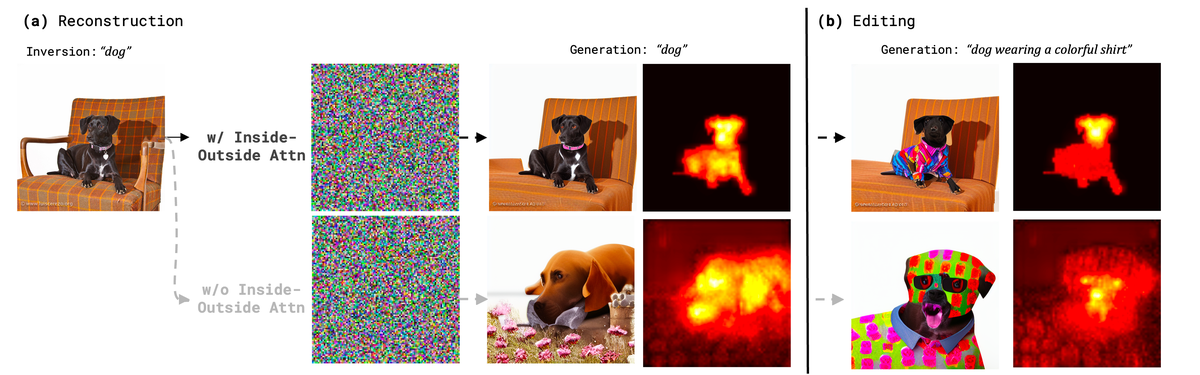}
\end{center}
\captionof{figure}{Spurious attentions and classifier-free guidance limits shape preservation.
Inside-Outside Attention (top) preserves the shape relationship between the object and background by associating tokens to specific spatial regions. We demonstrate this property when reconstructing (left) and editing (right) a real image with classifier-free guidance. We also depict the cross attention map for the token ``dog'' averaged all attention heads and timesteps.}
\label{fig:intuition}
\end{scriptsize}
\end{figure*}

\noindent\textbf{Structure Preserving Image Editing} Various works in image-to-image translation can also preserve structure during editing. To accomplish this, some works copy random seeds~\cite{wu2022unifying}, finetune model weights~\cite{valevski2022unitune, kawar2023imagic}, copy attention maps~\cite{hertz2022prompt}, or condition on a partially noised version of the source image ~\cite{meng2022sdedit}. There also exist a few concurrent works~\cite{mokady2022nulltext, tumanyan2022pnp, brooks2022instructpix2pix} that were developed independently and at the same time as our work, where we display a conceptual comparison in~\autoref{fig:concurrent-comparison} with additional examples in the Supplemental.
(a) While these methods are often able to mimic the general style and structure of the source image, they struggle to perform a local edit where the background is left undisturbed.
(b) This phenomenon occurs because these methods solely rely on the textual grounding capabilities of the diffusion model and~\textit{copy} attention maps, which are often noisy and entangle object and background pixels (see ~\autoref{fig:intuition}). On the other hand, we~\textit{constrain} these attention maps according to a shape, which can be derived from a user or from a more reliable automatic grounding module such as a segmentation model, thereby incorporating more accurate spatial localization in a training-free fashion. 
(c) Due to this entanglement, works such as Prompt-to-Prompt (P2P)~\cite{hertz2022prompt} often significantly drift when editing~\textit{real} images (see~\autoref{fig:intro}). Our Inside-Outside Attention mechanism mitigates this drift when applied to the inversion process (see~\autoref{fig:intuition}).\\
\noindent\textbf{Image Inpainting}
Image inpainting is the task of infilling the missing regions of an image.
Researchers have proposed dilated convolution~\cite{iizuka2017globally}, partial convolution~\cite{liu2018image}, gated convolution~\cite{yu2019free}, contextual attention~\cite{yu2018generative}, and co-modulation~\cite{zhao2021large} for GAN-based image inpainting. Lugmayr \etal \cite{lugmayr2022repaint} recently proposed a diffusion-based model for free-form image inpainting. There exist variants of GLIDE \cite{nichol2022glide} and Stable Diffusion \cite{rombach2021highresolution, sdinpaint} finetuned for text-conditional inpainting. However, these methods were trained with free-form masks without semantic meaning. There exist a few training-based methods that use object masks, none of which are publicly available. Make-a-Scene~\cite{gafni2022make} trained an auto-regressive transformer conditioned on full segmentation maps of a scene. Shape-guided Object Inpainting \cite{zeng2022shape} trained a GAN and Imagen Editor ~\cite{su2023imageneditor} trained Imagen \cite{saharia2022imagen} with object masks for inpainting. In contrast, we apply our model on top of an open-source text-to-image diffusion model at inference time. Because our method is training-free, it is more flexible and can be applied towards tasks beyond object editing, such as background editing or simultaneous inside-outside editing, as discussed in~\cref{subsec:additional}.

\section{\label{sec:approach} Shape-Guided Diffusion}
We present Shape-Guided Diffusion, a \textit{training-free} method that enables a pretrained text-to-image diffusion model to respect shape guidance. Our goal is to locally edit image $x_{src}$ given text prompts $ \mathcal{P}_{src}$ and $ \mathcal{P}_{edit}$ and optional object mask $m$ (inferred from $\mathcal{P}_{src}$ if not provided), so that edited image $x_{edit}$ is faithful to both $\mathcal{P}_{edit}$ and $m$. We introduce Inside-Outside Attention to explicitly constrain the cross- and self-attention maps during both the inversion (image to noise) and generation (noise to image) processes. An overview of our method can be found in~\autoref{fig:pipeline} and Alg.~\autoref{alg:sgd}. We build upon Stable Diffusion (SD), a Latent Diffusion Model (LDM)~\cite{rombach2021highresolution} that operates in low-resolution latent space. LDM latent space is a perceptually equivalent downsampled version of image space, meaning we are able to apply Inside-Outside Attention in latent space via downsampled object masks. For the rest of this paper, when we denote ``pixel'', ``image'', or ``noise'', we are referring to these concepts in LDM latent space.

\begin{algorithm}[H]
\caption{Shape-Guided Diffusion}\label{alg:shape-guided-diffusion-editing}
\textbf{Input:} A diffusion model $DM$ with autoencoder $\mathcal{E}, \mathcal{D}$, real image $x_{src}$, a source prompt $\mathcal{P}_{src}$, an edit prompt $\mathcal{P}_{edit}$, and either a binary object mask $m$ or a shape inference function InferShape$(\cdot)$.\\
\textbf{Hyperparameters:} Classifier-free guidance scale $w_g$.\\
\textbf{Output:} An edited image $x_{edit}$ that differs from $x_{src}$ only within the mask region $m$.
\begin{algorithmic}[1]
\If{$m$ is not provided}
    \State $m \gets$ \small{InferShape}($x_{src}, \mathcal{P}_{src}$)
\EndIf
\State $[\bar{z}_0, ..., \bar{z}_T] \sim$ \small{InsideOutsideInversion}$(z | \mathcal{E}(x_{src}), \mathcal{P}_{src}, m, DM)$
\State $z_{T} \gets \bar{z}_{T}$
\ForAll{$t$ from $T$ to $1$} 
    \State \small{InsideOutsideAttention}($DM, \mathcal{P}_{edit}, m$)
    \State $z_{cond} \gets DM(z_{t}, \mathcal{P}_{edit})$
    \State $z_{uncond} \gets DM(z_{t}, \varnothing)$
     \State $z_{t-1} \gets z_{cond} + w_g * (z_{cond} - z_{uncond})$
     \State $z_{t-1} \gets z_{t-1} \odot m + \bar{z}_{t-1} \odot (1 - m)$
\EndFor \\
$x_{edit} \gets \mathcal{D}(z_{0})$
\end{algorithmic}
\label{alg:sgd}
\end{algorithm}
\vspace{-0.3cm}

\begin{figure}[t]
\begin{scriptsize}
\begin{center}
\includegraphics[width=\linewidth]{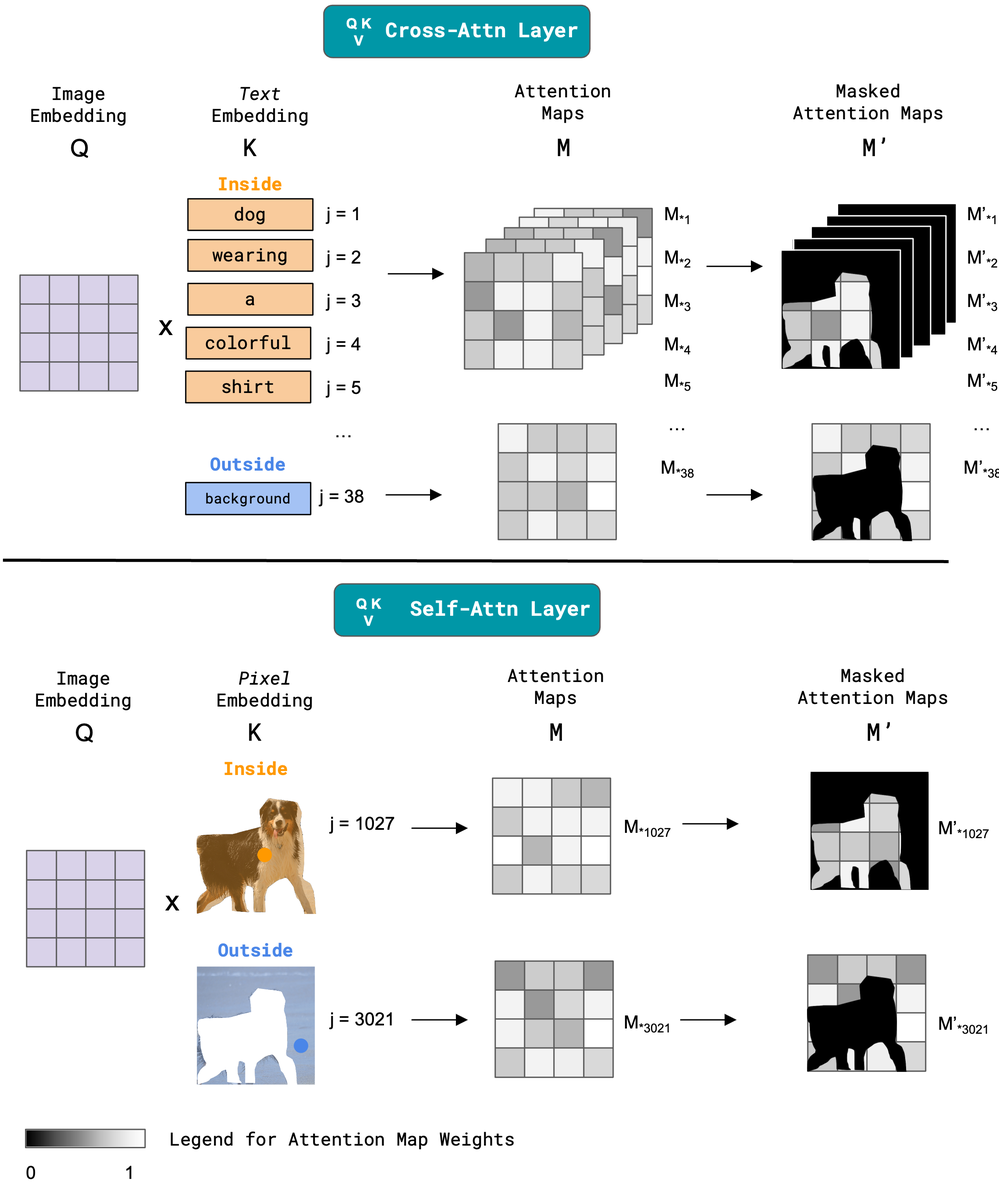}
\end{center}
\captionof{figure}{Inside-Outside Attention. We modify both the cross- and self-attention maps. Here $j$ refers to token/pixel indices and $M_{*j}$ denotes the attention map corresponding to the $j$-th index. Cross-Attn Layer (top): depending on whether the text embedding refers to the inside or outside the object, we constrain the attention map $M$ according to the object mask or the inverted object mask to produce $M'$. Self-Attn Layer (bottom): we perform a similar operation on the inside and outside pixel embeddings.}
\label{fig:inside-outside-attention}
\end{scriptsize}
\end{figure}

\subsection{Inside-Outside Attention}\label{subsec:inside_outside_attention}
LDMs contain both cross-attention layers used to produce a spatial attention map for each textual token and self-attention layers used to produce a spatial attention map for each pixel.
We postulate that prior methods often fail because of spurious attentions -- attentions that seek to edit the object compete with those that seek to preserve the background because they are not well localized (see~\autoref{fig:intuition}). Hence, we manipulate the cross-attention map such that the inside tokens are responsible for editing a distinct, non-overlapping spatial region compared with the outside tokens (e.g., ``dog'',``shirt'', etc. may only edit the dog and ``background'' may only edit the remaining scene). Since self-attention layers heavily influence how pixels are grouped to form coherent objects, we apply a similar manipulation to the self-attention map to further ensure that the desired object is contained within the boundaries of the input mask.\\
\indent An overview of Inside-Outside Attention is given in \autoref{fig:inside-outside-attention} and our algorithm is defined as follows (also see Alg.~\ref{alg:ioa}). 
For one forward pass at each timestep during inversion or generation, we go through all layers of the diffusion model $DM$ and manipulate the cross- and self-attention maps $M$. We denote the dimensions of $M$ as $\mathbb{R}^{HW \times d_{\tau}}$ and $\mathbb{R}^{HW \times HW}$ for each cross- and self-attention map, respectively, where $H$ is the image height, $W$ is the image width, $HW$ is the number of pixels in the flattened image, and $d_{\tau}$ is the number of tokens. We also downsample $m$ according to the resolution of the cross- or self-attention layer. For the cross-attention map, we determine column indices $J_{in}$ and $J_{out}$ based on whether the token refers to the object or the background. For the self-attention map, we determine column indices $J_{in}$ and $J_{out}$ based on whether the pixel belongs inside or outside the object as defined by mask $m$. Finally, we compute the new constrained attention maps $M_{*j_{in}}' = \{M_{*j_{in}} \odot m \mid \forall j_{in} \in J_{in}\}$ and $M_{*j_{out}}' = \{M_{*j_{out}} \odot (1 - m) \mid \forall j_{out} \in J_{out}\}$.

\begin{algorithm}[H]
\caption{Inside-Outside Attention}\label{alg:inside-outside-attention}
\textbf{Input:} A diffusion model $DM$, a binary object mask $m$, a prompt $\mathcal{P}$. \\
\textbf{Output:} An edited diffusion model where the attention maps $M$ are masked according to $m$ and $\mathcal{P}$ for one forward pass.
\begin{algorithmic}[1]
\ForAll{$l \in \text{layers}(DM)$}
    \State \textbf{if} type($l$) is CrossAttention
    \State \quad\quad $J_{in} \gets \{j \mid$ $j$th token refers to object$\}$
    \State \quad\quad $J_{out} \gets \{j \mid$ $j$th token refers to background$\}$
    \State \textbf{elif} type($l$) is SelfAttention
    \State \quad\quad $J_{in} \gets \{j \mid$ $j$th pixel belongs inside object$\}$
    \State \quad\quad $J_{out} \gets \{j \mid$ $j$th pixel belongs outside object$\}$
    \State $M_{*j_{in}}' = M_{*j_{in}} \odot m \quad \forall j_{in} \in J_{in}$
    \State $M_{*j_{out}}' = M_{*j_{out}} \odot (1 - m) \quad \forall j_{out} \in J_{out}$
\EndFor
\end{algorithmic}
\label{alg:ioa}
\end{algorithm}
\vspace{-0.2cm}

\subsection{Inside-Outside Inversion}\label{subsec:inversion}
To edit real images, we use DDIM inversion~\cite{song2020denoising, nichol2022glide} to convert the source image to inverted noise. However, we observe that using inversion with a text-to-image diffusion model often results in a shape-text faithfulness tradeoff. While running generation with the fully conditional or unconditional model can reconstruct the real image, using non-zero levels of classifier-free guidance can completely drift from the real image, as seen in the bottom row of~\autoref{fig:intuition}. We propose applying Inside-Outside Attention to mitigate this trade-off. Similar to how prior work can associate tokens to \textit{entire images}~\cite{gal2022image, valevski2022unitune}, with Inside-Outside Attention we can associate tokens to \textit{specific spatial regions}. As seen in the top row of ~\autoref{fig:intuition}, applying our mechanism during inversion and generation allows one to both reconstruct and edit the real image with classifier-free guidance. We also visualize the cross-attention maps for the token ``dog,'' where with our mechanism its effect is constrained to the silhouette and leaves the chair in the background unaffected, whereas without our mechanism its effect leaks into the background and morphs the dog while removing the chair.

\subsection{Method Summary}\label{subsec:generation}
In summary, we make the observation that object shape can be better preserved if spurious attentions are removed, and we propose the novel inference-time mechanism Inside-Outside Attention. Our method Shape-Guided Diffusion uses Inside-Outside Attention to constrain the attention maps during both inversion and generation, which we depict in~\autoref{fig:pipeline}. The  Shape-Guided Diffusion algorithm can be defined as follows (also see Alg.~\ref{alg:sgd}). If the mask is not provided, we use the shape inference function InferShape$(\cdot)$ to identify $\mathcal{P}_{src}$ in the image. For our experiments we use an off-the-shelf segmentation model~\cite{cheng2021maskformer}, but any method for textual grounding could also be used with our method. We run Inside-Outside Inversion on the conditional diffusion model driven by the prompt $\mathcal{P}_{src}$  (e.g., ``dog'') to get inverted noise $\bar{z}_{T}$. We then set our initial noise $z_{T}$ to $\bar{z}_{T}$. For each sampling step, we apply Inside-Outside Attention for both the conditional and unconditional diffusion models using mask $m$ and $\mathcal{P}_{edit}$ (e.g., ``dog wearing a colorful shirt''). We mix the predictions of both models using the original formulation of Ho \etal~\cite{ho2020denoising}, which applies classifier-free guidance to the conditional prediction (Line 10, Alg.~\ref{alg:sgd}). In early experiments we found this design choice leads to higher text alignment without a loss in other metrics. Finally, we copy the real image's background found during the inversion process $\bar{z}_{t-1} \cdot m$ to form the edited image prediction $z_{t-1}$. This ensures the edited image $x_{edit}$ and the original image $x_{src}$ only differ within the mask region $m$.

\begin{figure*}
\vspace{-0.5cm}
\begin{scriptsize}
\begin{center}
\includegraphics[width=0.88\linewidth]{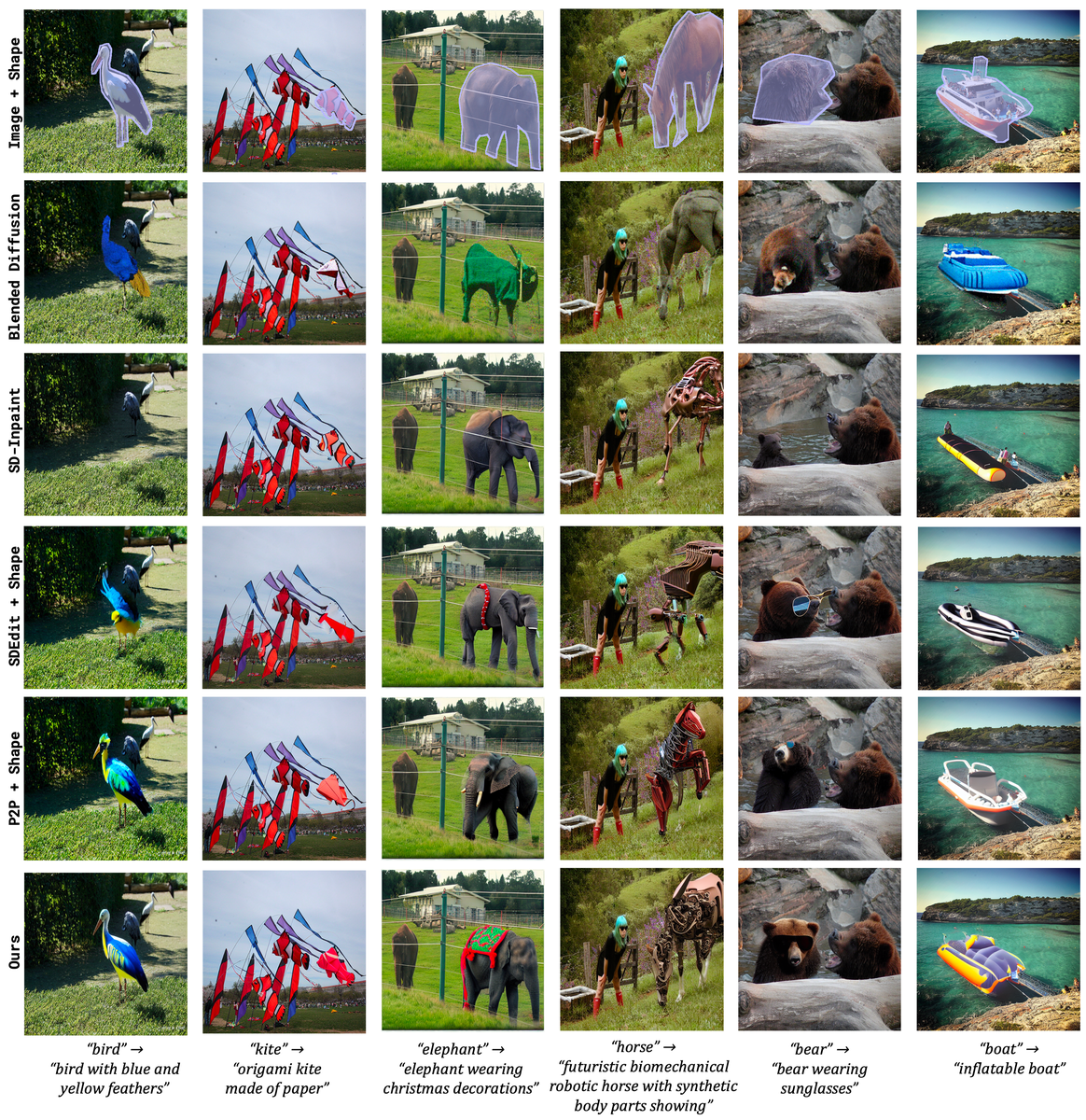}
\end{center}
\captionof{figure}{Comparison to prior work. We compare our results with Blended Diffusion~\cite{avrahami2022blended}, SD-Inpaint~\cite{sdinpaint}, SDEdit~\cite{meng2022sdedit}, and P2P~\cite{hertz2022prompt} on MS-COCO images.
Our method is able to generate realistic edits that are faithful to both the input shape and text prompt.
+ Shape denotes a variant of the structure preserving method adapted for local image editing using the ``copy background'' method from \cite{avrahami2022blended}.}
\label{fig:comparisons}
\end{scriptsize}
\end{figure*}

\section{\label{sec:evaluation} MS-COCO ShapePrompts}
\paragraph{Benchmark} We evaluate our approach on MS-COCO images~\cite{lin2014coco}.
We filter for object masks with an area between [2\%, 50\%] of the image, following prior work in image inpainting \cite{suvorov2021resolution}.
Our test set derived from MS-COCO val 2017 contains $1,149$ object masks spanning $10$ categories covering animal, vehicle, food, and sports classes.
We create a validation set with $1,000$ object masks in the same fashion derived from MS-COCO train 2017. For each category we design a few prompts that add clothing or accessories (e.g., ``floral shirt'' or ``sunglasses''), manipulate color (e.g., ``iridescent'', ``with spray paint graffiti''), switch material (``lego'', ``paper''), or specify rare subcategories (``spotted leopard cat'', ``tortilla wrapped sandwich''). More information about the prompts can be found in the Supplemental.\\
\noindent\textbf{Metrics} Since we aim to synthesize an image faithful to the input shape, we use mean Intersection over Union (mIoU) as a metric. Specifically, we compute the proportion of pixels within the masked region correctly synthesized as the desired object class,
as determined by a segmentation model~\cite{cheng2021maskformer} trained on COCO-Stuff~\cite{caesar2018cocostuff}.
Since animal object masks are particularly fine-grained, and mIoU does not capture a full picture of degenerate cases (e.g., if the edit replaces a cat's full body with a cat's head), we also compute a keypoint-weighted mIoU (KW-mIoU) for the animal classes. Specifically, we weight each sample's mIoU by the percentage of correct keypoints when comparing the source vs. edited image, as determined by an animal keypoint detection model~\cite{yu2021ap}. We also report FID scores as a metric for image realism, which measures the similarity of the distributions of real and synthetic images using the features of an Inception network~\cite{szegedy2016rethinking, parmar2021cleanfid}. 
Finally, we report CLIP~\cite{radford2021learning} scores as a metric for image-text alignment, which measures the similarity of the text prompt and synthetic image using the features of a large pretrained image-text model. More information on metrics can be found in the Supplemental.

\begin{figure*}
\vspace{-0.3cm}
\begin{scriptsize}
\begin{center}
\includegraphics[width=\linewidth]{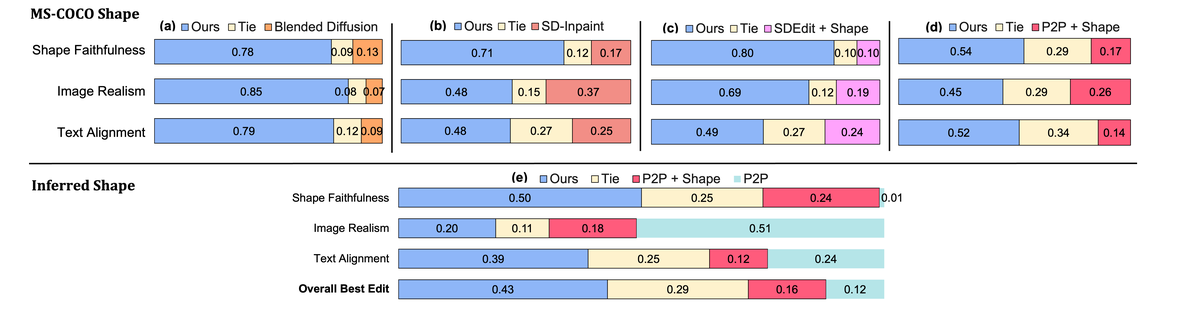}
\end{center}
\captionof{figure}{Annotator evaluation on MS-COCO ShapePrompts (100-sample subset of test set). Columns (a, b, c, d): we asked people to rate edits performed by our method vs. a baseline, where the two edits were presented as anonymized and in randomized order. Rows (shape faithfulness, image realism, text alignment): annotators selected the superior edit along these three axes. Each bar denotes the percentage of samples where the superior edit was ``Ours'', ``Tie'', or a baseline. In (e) we use the same procedure, except we presented three anonymized edits, ours vs. two baselines. Annotators were additionally asked to select the ``overall best edit.'' We provide further details in the Supplemental.
}
\label{fig:human_test100}
\end{scriptsize}
\end{figure*}
\section{\label{sec:experiments} Experiments}
In \cref{subsec:comparisons} we evaluate our method on the shape-guided editing task where it must replace an object given a (real image, text prompt, object mask) triplet from MS-COCO ShapePrompts. We also evaluate on the same task with masks inferred from the text and ablate the use of our Inside-Outside Attention mechanism. In \cref{subsec:additional} we present additional results beyond object editing.

\noindent\textbf{Baselines} For our baselines, we compare against the local image editing method Blended Diffusion~\cite{avrahami2022blended}, the inpainting method SD-Inpaint~\cite{sdinpaint}, and the structure preserving methods SDEdit~\cite{meng2022sdedit} and P2P~\cite{hertz2022prompt}. Blended Diffusion, built on top of a Guided Diffusion~\cite{dhariwal2021guided} backbone, uses mask input by copying the source image's background at each timestep and text input by applying classifier guidance with CLIP~\cite{radford2021learning}. SD-Inpaint, built on top of a Stable Diffusion~\cite{rombach2021highresolution} backbone, finetunes the model with an extra U-Net channel to use mask input and applies classifier-free guidance to use text input. SDEdit partially noises then denoises the source image and P2P copies cross attention maps to preserve structure, and they apply classifier-free guidance to use text input. For the structure preserving methods we use implementations built on top of a Stable Diffusion backbone, and in some experiments we adapt them to use mask input by applying the ``copy background'' method from~\cite{avrahami2022blended}.

\begin{table}
\begin{center}
\begin{scriptsize}
\begin{tabular}{lllll}
\toprule
Approach & KW-mIoU $(\uparrow)$ & mIoU $(\uparrow)$ & FID $(\downarrow)$ & CLIP $(\uparrow)$\\
\midrule
Real Images & 83.3 & 76.3 & - & 0.15 \\
\hline & \\[-0.2cm]
\textbf{MS-COCO Shape} \\
Blended Diffusion \cite{avrahami2022blended} & 23.3 & 41.8 & 46.2 & 0.20 \\
SD-Inpaint \cite{sdinpaint} & 38.5 & 51.7 & 43.7 & 0.19 \\
SDEdit + Shape~\cite{meng2022sdedit} & 31.0 & 49.9 & 45.1 & \textbf{0.21}\\
P2P + Shape~\cite{hertz2022prompt} & 46.9 & 63.3 & \textbf{39.6} & 0.20\\
Ours (w/o IOA) & 43.8 & 55.3 & 41.5 & 0.21\\
Ours & \textbf{53.3} & \textbf{63.6} & 40.2 & \textbf{0.21}\\
\hline & \\[-0.2cm]
\textbf{Inferred Shape} \\
P2P \cite{hertz2022prompt} & 24.2 & 64.6 & 97.5 & \textbf{0.26}\\
P2P + Shape \cite{hertz2022prompt} & 37.7 & 54.0 & 51.1 & 0.21\\
Ours (w/o IOA) & 33.0 & 46.0 & 56.8 & 0.22\\
Ours & \textbf{43.0} & \textbf{54.9} & \textbf{49.5} & 0.22\\
\bottomrule
\end{tabular}
\end{scriptsize}
\end{center}
\captionof{table}{Automatic evaluation on MS-COCO ShapePrompts (test set). MS-COCO Shape uses object masks provided by MS-COCO, and Inferred Shape uses object masks inferred from the text. Ours w/o IOA denotes our method without Inside-Outside Attention.}
\label{tab:test}
\end{table}

\begin{figure}[t]
\begin{scriptsize}
\begin{center}
\includegraphics[width=\linewidth]{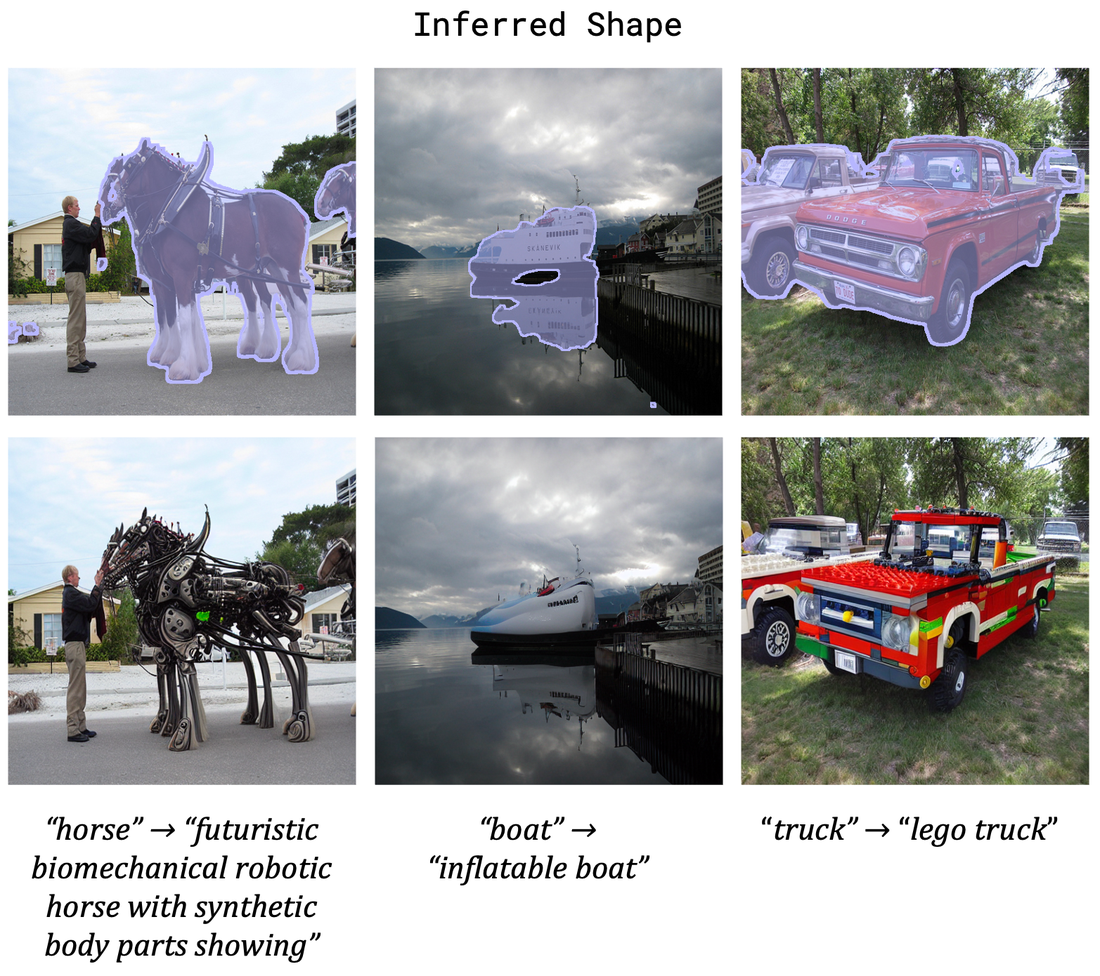}
\end{center}
\captionof{figure}{Our method can handle challenging cases present in automatically inferred object masks such as overlapping instances or reflection.
}
\label{fig:automatic-mask}
\end{scriptsize}
\end{figure}

\noindent\textbf{Experimental Setup} For all baselines we use the default hyperparameters provided by their respective repositories. For sampling we use a standard DDIM scheduler for $50$ inversion and generation steps. When using Inside-Outside Attention on cross-attention layers, we evenly divide the maximum number of text tokens excluding the \texttt{<bos>} token, resulting in $38$ ``inside'' tokens and $38$ ``outside'' tokens. The attentions for the \texttt{<bos>} token are zeroed out.

\subsection{Comparison to Prior Work}\label{subsec:comparisons}
\noindent\textbf{MS-COCO Shape} We first experiment with object masks provided by MS-COCO as our shape guidance. In \autoref{fig:comparisons}, we depict real images (first row) and edits made by Blended Diffusion (second row), SD-Inpaint (third row), and SDEdit + Shape (fourth row), P2P + Shape (fifth row), Ours (sixth row). Prior works demonstrate a variety of failure modes in shape-guided editing, where an object may be transformed into a new shape, removed completely, severely downscaled, or fail to respect the text prompt. On the other hand, our method is able to simultaneously respect the shape and the prompt without a compromise in image realism. As seen in~\autoref{tab:test}, our method outperforms the local editing and inpainting baselines \cite{avrahami2022blended, sdinpaint} across the board, with at least a 15 point improvement in KW-mIoU.
Comparing with the structure preserving baselines \cite{meng2022sdedit, hertz2022prompt}, we achieve at least a 6 point improvement in KW-mIoU with comparable FID and CLIP scores. We hypothesize that our method is able to significantly outperform the baselines in shape faithfulness because our constraint operates on attention maps, whereas ``copy background'' operates on model outputs, which can be less meaningful at early timesteps when outputs are close to pure noise. See the Supplemental for further discussion.\\
\indent We also conducted an evaluation with annotator ratings. We created four evaluations corresponding to each baseline, each of which contained 100 samples comparing an edit made by our method vs. the baseline in an anonymized and randomized fashion. For each sample, we asked five people to select the superior edit along the axes of shape faithfulness, image realism, and text alignment.
As seen in the top row of \autoref{fig:human_test100}, annotators confirm that our method outperforms the baselines in shape faithfulness, with our method selected as superior at least 54\% of the time (3.2x the most competitive baseline P2P + Shape). For image realism and text alignment, our method was selected as superior at least 48\% of the time (1.3x and 1.9x the most competitive baseline SD-Inpaint).

\begin{figure}[t]
\begin{scriptsize}
\begin{center}
\includegraphics[width=\linewidth]{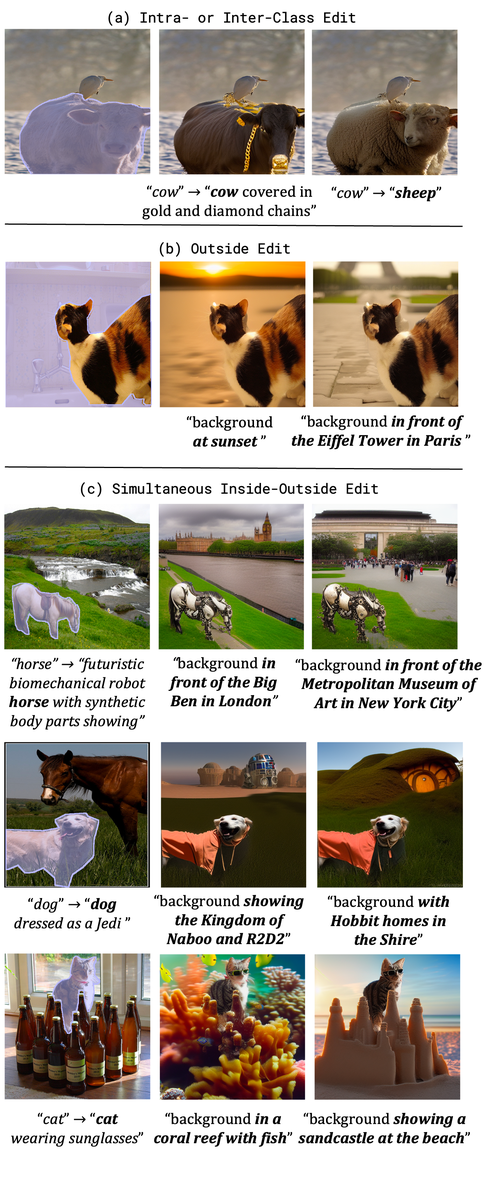}
\end{center}
\captionof{figure}{Additional editing results. Our method can perform intra- or inter-class edits on the same image, outside edits, and simultaneous inside-outside edits.}
\label{fig:inter-class-edit}
\end{scriptsize}
\end{figure}

\noindent\textbf{Inferred Shape} Next, we demonstrate that our method also works on automatically inferred masks (\autoref{fig:automatic-mask}). We compare against our most competitive baseline, vanilla P2P and P2P adapted for local image editing using the inferred mask (P2P + Shape). P2P often produces edits that look nothing like the source image (see~\autoref{fig:intro}), so annotators rate it as the worst overall image \textit{editing} method (see ~\autoref{fig:human_test100}). In fact, the method often produces simple images with a prominent single object, which significantly deviates from the distribution of complex, multi-object MS-COCO images. As a result, as seen in~\autoref{tab:test} these types of images achieve the worst FID scores with unusually high CLIP scores because these types of images tend to maximize text-image alignment scores, at the cost of faithfully preserving content in the real image. In contrast, our method is rated by annotators as the best image editing method for 43\% of samples, 2.7x more than the most competitive baseline P2P + Shape (see ~\autoref{fig:human_test100}). We also outperform P2P + Shape in all automatic metrics in~\autoref{tab:test}.\\
\noindent\textbf{Ablations} In~\autoref{tab:test} we ablate our Inside-Outside Attention mechanism (Ours w/o IOA vs. Ours).
The mechanism is a critical component of our method, providing a 9.5 point and 10 point increase in KW-mIoU in the MS-COCO Shape and Inferred Shape settings respectively. Ours w/o IOA performs better than all baselines on all metrics, except P2P + Shape (only P2P and our method use inversion), demonstrating how DDIM inversion is another critical component. In the Supplemental we also ablate the effect of DDIM inversion, guidance scale hyperparameters, and a soft vs. hard shape constraint on the self-attention maps.

\subsection{Additional Editing Results}\label{subsec:additional} In \autoref{fig:inter-class-edit}, we demonstrate additional capabilities of our method beyond object editing. 
(a) Our method is able to perform both intra- and inter- class edits on the same image, including adding accessories to a cow or transforming it into a sheep.
(b) Our method is able to perform outside edits, including changing the time of day or location. Because we invert the image prior to editing, our method sometimes maintains structures from the real image, for example transforming the cabinet into a landmass in both edited images.
(c) Our method is able to perform simultaneous edits with one prompt for the inside region and another for the outside region. 
Since our method delineates edits on the object vs. background, although every pixel in the image is transformed we can maintain the object-background relation from the source scene.
In contrast, it is not obvious how to adapt structure preserving methods for this simultaneous editing setting, since with ``copy background'' they require one region (e.g., the background) to remain identical to the source image to enforce locality.

\section{\label{sec:conclusion} Conclusion}
In this work, we present Shape-Guided Diffusion, a training-free method for utilizing precise object silhouette as a constraint in text-to-image diffusion models, which can be user provided or automatically inferred from the text. While prior work fails to respect shape inputs, our novel Inside-Outside Attention mechanism removes spurious attentions and localizes object vs. edits. We evaluate our method on our newly proposed MS-COCO ShapePrompts benchmark on the shape-guided editing task, where the goal is to edit an object given an input mask and text prompt. We show that our method significantly outperforms the baselines in shape faithfulness without a degradation in text alignment or image realism.

{\small
\bibliographystyle{ieee_fullname}
\bibliography{egbib}
}

\clearpage
\pagebreak
\section*{\label{sec:appendix} Supplementary Material}
In~\cref{sec:concurrent} we include a qualitative comparison with concurrent structure preserving methods, demonstrating how these methods often edit objects unmentioned by the text prompt. In~\cref{sec:shape_prompts_details} we discuss details regarding the MS-COCO ShapePrompts benchmark, in~\cref{sec:qual_eval_details} we discuss details regarding our annotator evaluation, and in~\cref{sec:ablations} we report additional ablations. Finally, we show a variety of additional examples from our method including success and failure cases in~\cref{sec:success_failure} as well as inferred shape edits, inter-class edits, and outside edits in~\cref{sec:additional_settings}. 

\section{\label{sec:concurrent}Qualitative Comparison with Concurrent Structure Preserving Methods}
We compare our method with concurrent work~\cite{mokady2022nulltext, tumanyan2022pnp, brooks2022instructpix2pix}. We can see that our method is able to perform better localized edits on a real image. Because these methods lack an explicit shape, they often change irrelevant objects that are not specified in the text prompt. In Row 2, Col 1-3 not only is the horse transformed into a robot but also the man. 
In Row 5, Col 3, 5 the wall that was present in the real image disappears. In contrast, the variant of our method that uses automatically inferred shapes (thereby requiring the same amount of input as the structure preserving methods) is able to perform edits that only modify the object of interest without disturbing the background. We used the official codebases released by the respective baselines and generated results using their default hyperparameter settings.

\section{\label{sec:shape_prompts_details} MS-COCO ShapePrompts Details}
\begin{figure}[h]
\begin{scriptsize}
\begin{center}
\includegraphics[width=\linewidth]{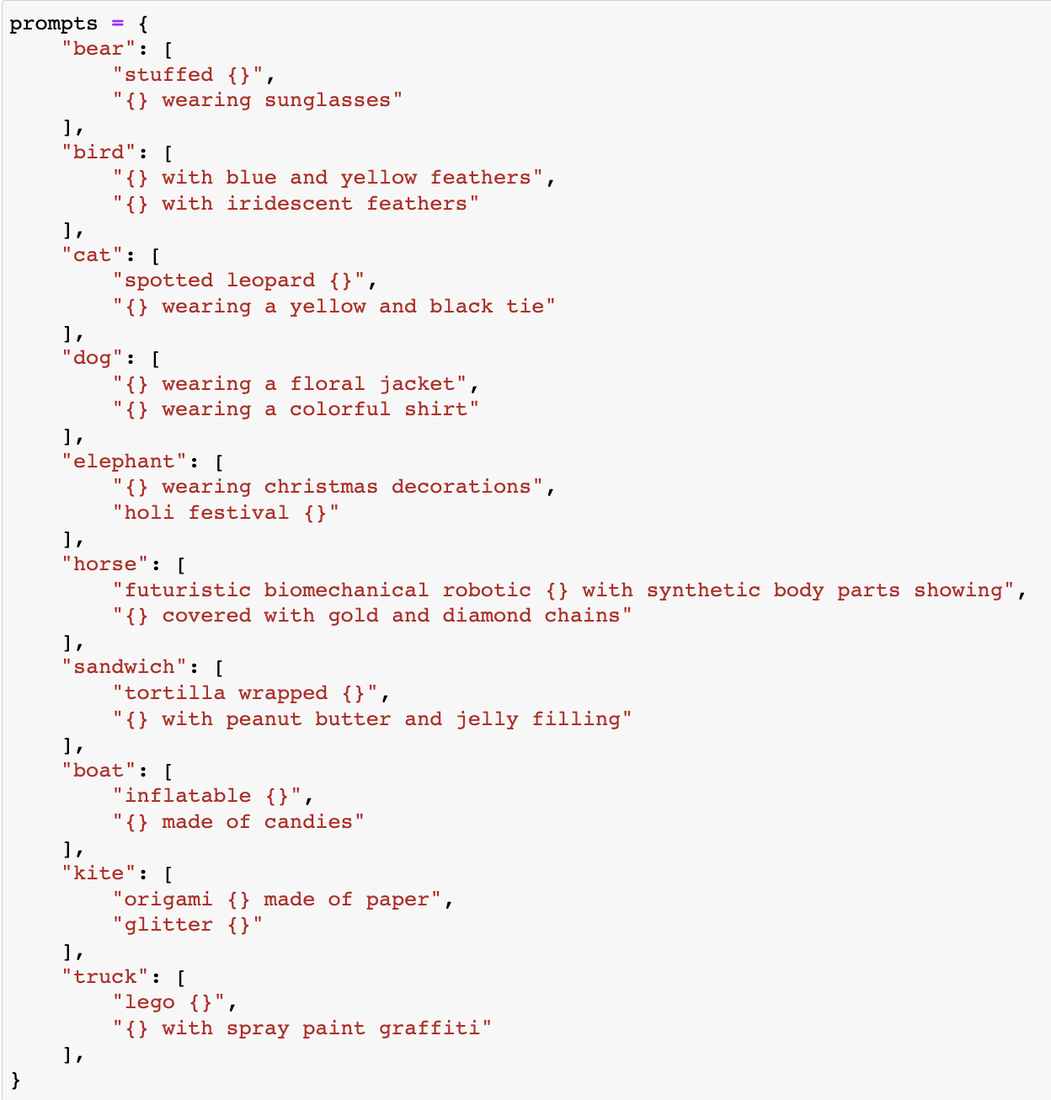}
\end{center}
\caption{Prompts from the MS-COCO ShapePrompts benchmark.}
\label{fig:prompts}
\end{scriptsize}
\end{figure}

\noindent \textbf{Prompts} For our MS-COCO ShapePrompts benchmark we design a set of prompts where it is possible to simultaneously synthesize an object that is shape faithful and text aligned (as opposed to prompts entangled with shape, i.e., transforming ``chihuahua dog'' to ``poodle dog'' while respecting shape is difficult because poodles are characterized by floppy ears and fluffy fur). While our method is able to perform inter-class edits as seen in \autoref{fig:additional_inter_class}, we focus our experiments on intra-class edits which make more sense given the shape constraint (e.g. some hyper-specific shapes like the silhouette of an elephant only make sense when edits are done within the object class). For this reason we design prompts for each object class as seen in \autoref{fig:prompts}. These prompts were inspired by examples from prior work~\cite{avrahami2022blended, hertz2022prompt} and a search engine with paired prompts and synthetic images from Stable Diffusion~\cite{lexica}.

\begin{figure}
\begin{scriptsize}
\begin{center}
\includegraphics[width=\linewidth]{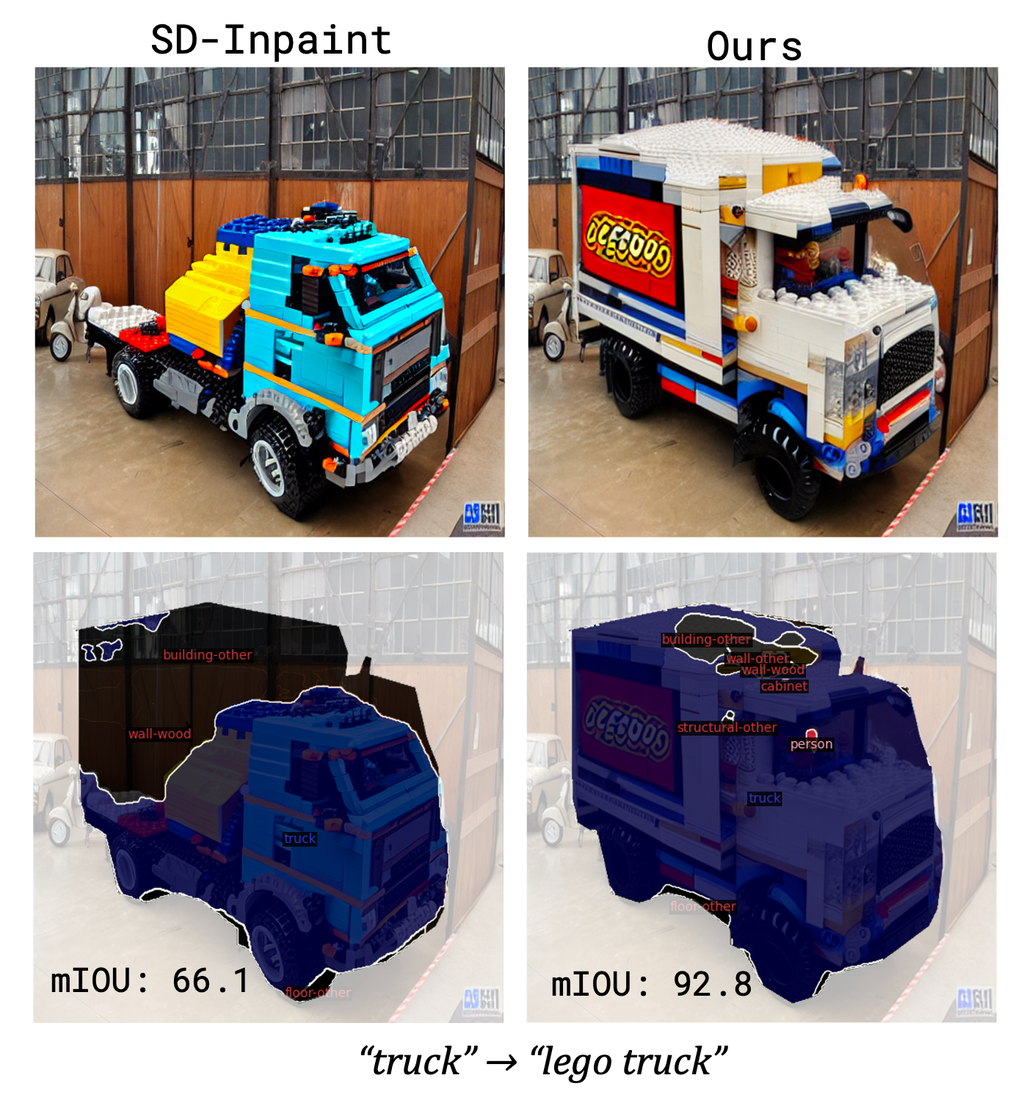}
\end{center}
\caption{Synthetic images and their corresponding predicted segmentation and mIoU. Out-of-distribution variants of the object class, such as a truck made of legos, are still segmented correctly.
}
\label{fig:segmentation}
\end{scriptsize}
\end{figure}

\begin{figure}[t]
\begin{scriptsize}
\begin{center}
\includegraphics[width=\linewidth]{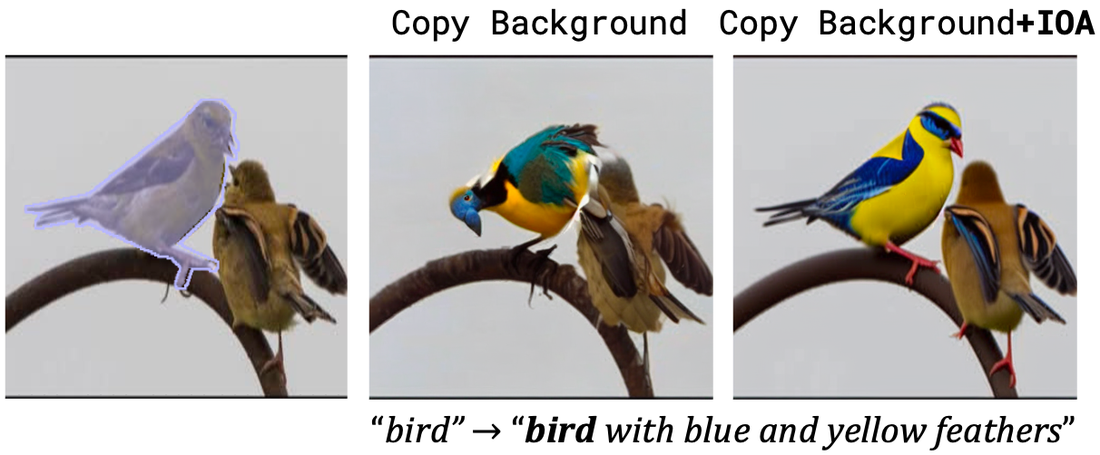}
\end{center}
\captionof{figure}{Shape signal from ``copy background'' is weak in early timesteps. In both examples we only use shape guidance in the first half of generation, where Inside-Outside Attention (+IOA) is able to provide stronger shape signal.}
\label{fig:copy-background}
\end{scriptsize}
\end{figure}

\begin{table*}[t]
\begin{center}
\begin{small}
\begin{tabular}{l|l|llll}
\toprule
Approach & Guidance Scale & KW-mIoU & mIoU $(\uparrow)$ & FID $(\downarrow)$ & CLIP $(\uparrow)$\\
\midrule
Real Images & N/A & 86.5 & 78.6 & - & 0.16\\
\midrule
(1) SD & 7.5 & 30.9 & 52.5 & 46.2 & 0.21\\
(2) SD + DDIM Inv & 7.5 & 39.8 & 61.2 & 42.8 & 0.21\\
(3) SD + DDIM Inv + Re-Weight (Ours w/o IOA) & 3.5 & 46.2 & 59.6 & 40.6 & 0.21\\
(4) SD + DDIM Inv + Re-Weight + Token Inside-Outside Attn & 3.5 & 48.1 & 62.9 & 40.6 & 0.21\\
(5) SD + DDIM Inv + Re-Weight + Soft Inside-Outside Attn & 3.5 & 51.4 & 66.2 & 40.2 & 0.21\\
(6) SD + DDIM Inv + Re-Weight + Hard Inside-Outside Attn (Ours) & 3.5 & \textbf{54.8} & \textbf{67.6} & \textbf{39.0} & 0.21\\
\bottomrule
\end{tabular}
\end{small}
\end{center}
\caption{Ablations on MS-COCO ShapePrompts (validation set).}
\label{tab:ablation_val}
\end{table*}

\begin{figure}
\begin{scriptsize}
\begin{center}
\includegraphics[width=\linewidth]{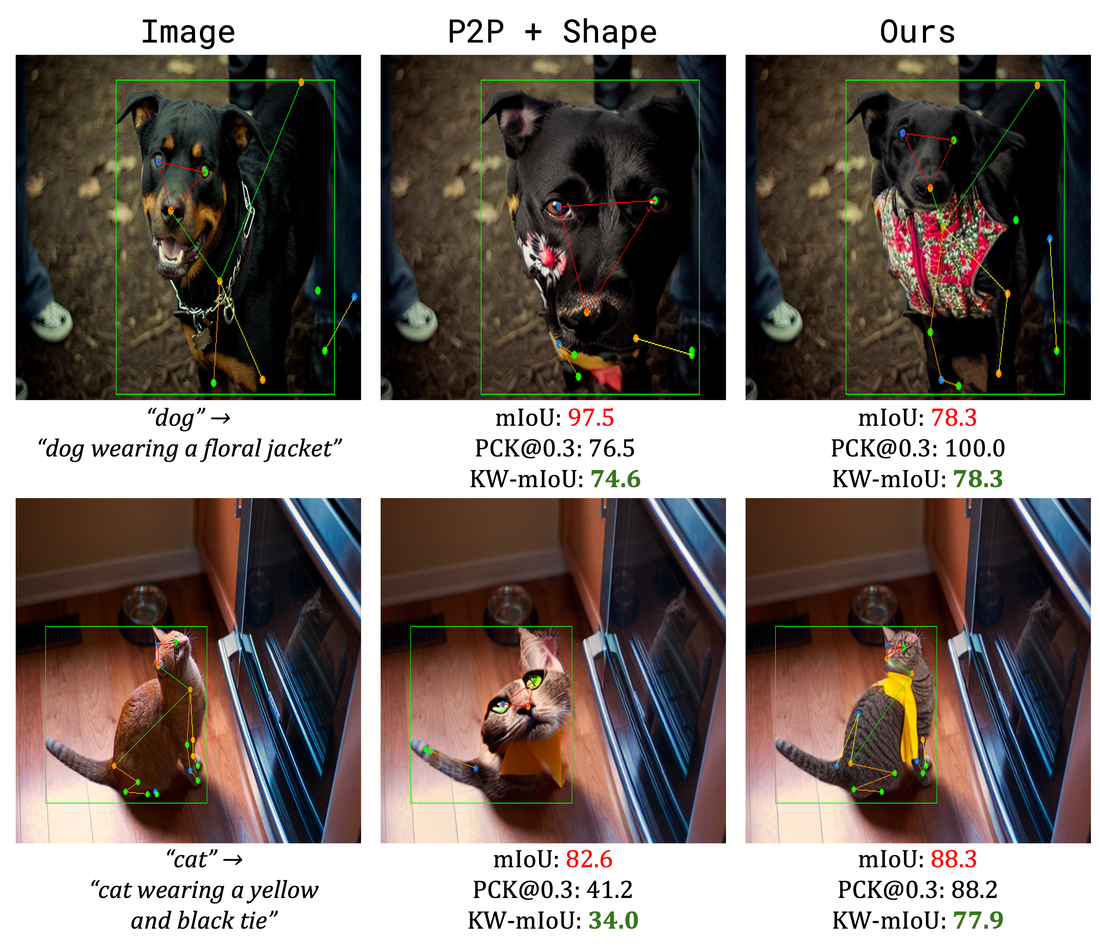}
\end{center}
\caption{Comparison between mIoU and Keypoint-Weighted mIoU (KW-mIoU). Note that in this example P2P~\cite{hertz2022prompt} receives a high mIoU score even though the edited objects are incorrectly scaled or cut off. By weighting each sample's mIoU with the percentage of correct keypoints (PCK) to compute the KW-mIoU, we can measure shape faithfulness more reliably.}
\label{fig:kw-miou}
\end{scriptsize}
\end{figure}

\begin{figure}[t]
\begin{scriptsize}
\begin{center}
\includegraphics[width=\linewidth]{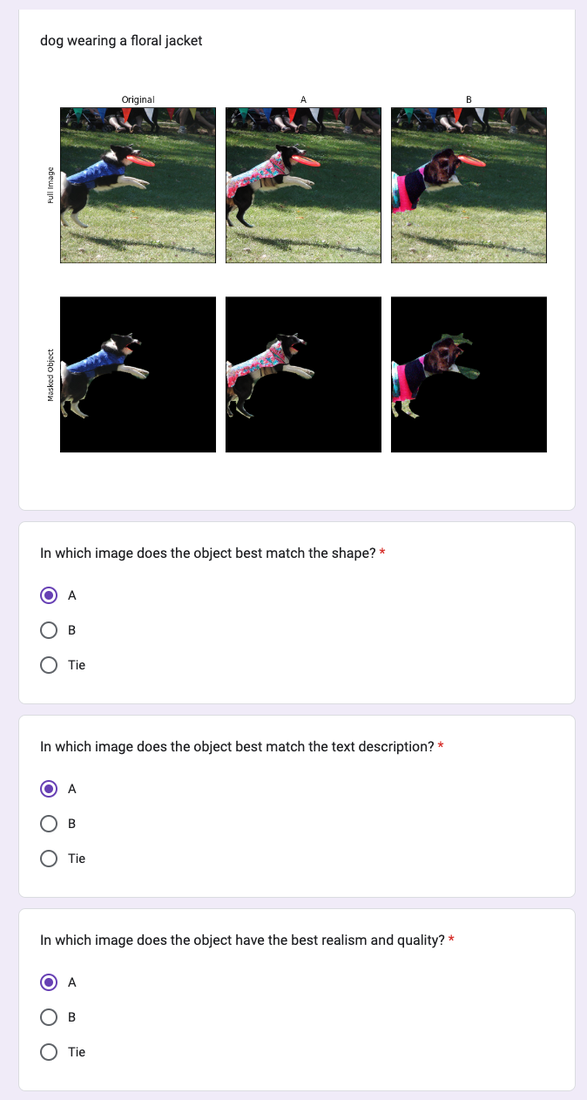}
\end{center}
\caption{Screenshot of our annotator evaluation. People were asked to compare images edited by our method versus a baseline in anonymized and randomized order and rate (1) shape faithfulness, (2) text alignment, (3) image realism. In our final evaluation comparing Ours vs. P2P vs. P2P + Shape they also rated (4) best overall edit.}
\label{fig:google-form}
\end{scriptsize}
\end{figure}

\begin{figure}[t]
\begin{scriptsize}
\begin{center}
\includegraphics[width=\linewidth]{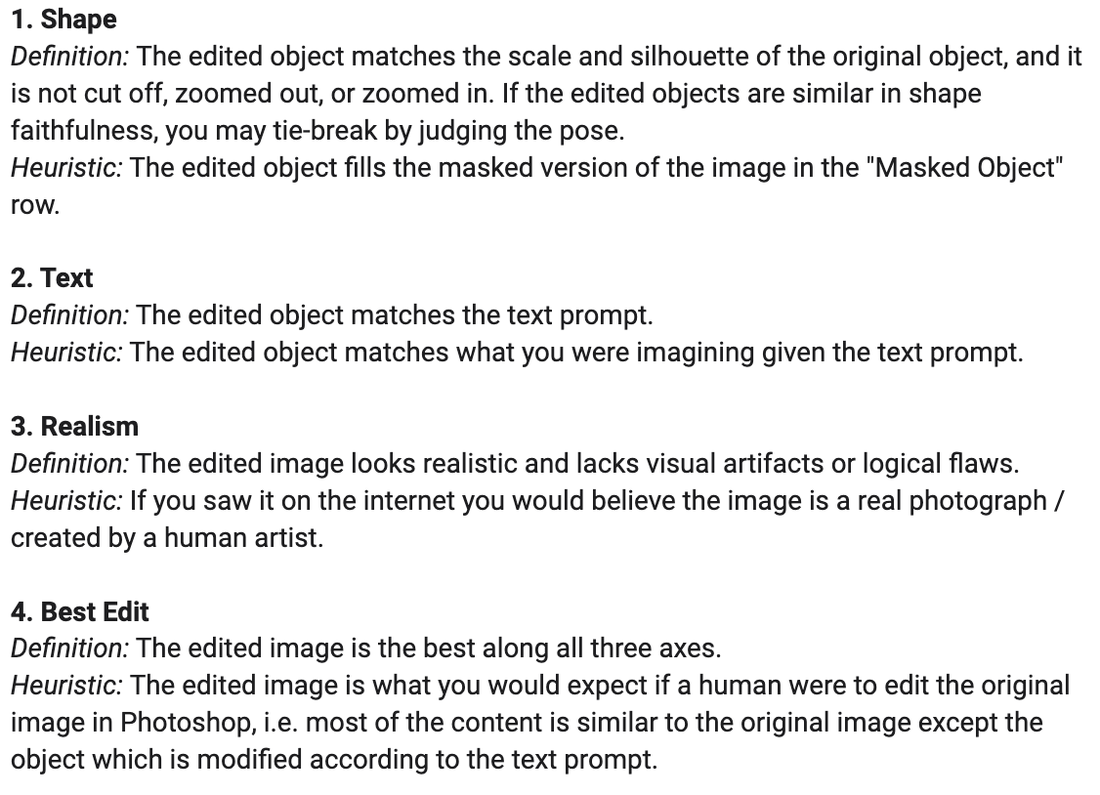}
\end{center}
\caption{Metric definitions given as reference in the annotator evaluation.}
\label{fig:metric-definitions}
\end{scriptsize}
\end{figure}

\noindent \textbf{Shape Faithfulness Metric} To measure shape faithfulness we use the pretrained segmentation model MaskFormer \cite{cheng2021maskformer}. We demonstrate that the model makes meaningful predictions on synthetic images in \autoref{fig:segmentation}. Even more, the model's predictions are reasonably robust to out-of-distribution variants of the object class, such as ``lego truck.'' 
We use the segmentation model to compute mean intersection over union (mIoU). We compute mIoU on a per-sample basis (i.e., we average the IOU of each object regardless of size) as opposed to a per-pixel basis (which is typically used in semantic segmentation works) since it is equally important to synthesize both small and large objects in a shape-faithful fashion. We set all pixels outside the mask to a null prediction to compute mIoU only within the edited mask region. We do this because in some settings (e.g. MS-COCO instance masks) the mask may specify one object instance out of multiple to edit, but our segmentation model would identify all instances of the same category, which would result in a diluted mIoU score. Additionally, for all methods that use ``copy background'' the background should remain identical to the original image.\\
\indent In addition to mIoU, we introduce a new metric called Keypoint-Weighted mIoU (KW-mIoU). One issue with the standard mIoU metric is that edited objects that are incorrectly scaled or cut off could still get a very high mIoU if they fully occupy the shape (see \autoref{fig:kw-miou}, Col 2). In order to mitigate this issue, we report KW-mIoU for animal classes (horse, dog, cat, elephant) where we weight each sample's mIoU by the percentage of correct keypoints (PCK) as computed between the source and edited images. We report KW-mIoU for animal classes only as we were not able to find a robust object pose estimation model with open-vocabulary capacities and reliable performance. By incorporating pose information, the proposed metric is able to be more sensitive to scale and object parts, and thus measure shape faithfulness more reliably. We use an animal keypoint detection model HRNet~\cite{sun2019deep} pretrained on AP-10K dataset~\cite{yu2021ap} as provided by \url{https://github.com/open-mmlab/mmpose}.

\section{\label{sec:qual_eval_details} Annotator Evaluation Details}
Our annotator evaluation included 25 total people spread across 5 evaluations (Ours vs. Blended Diffusion, Ours vs. SD-Inpaint, Ours vs. SDEdit + Shape, Ours vs. P2P + Shape, Ours vs. P2P vs. P2P + Shape). We asked each annotator to rate 100 samples, where they were told that they would be ``rating AI-edited images, where the goal is to edit one object according to a text prompt while maintaining its shape.'' Each sample was formatted as pictured in \autoref{fig:google-form}, in the grid the first column (``Original'') corresponds to the original image, the second column (``A'') corresponds to an edited image, and the third column (``B'') corresponds to another edited image. To help annotators judge faithfulness in addition to the first row (``Full Image'') we also provide the second row (``Masked Object'') which masks the full image according to the shape of the original object. Along the metrics of shape faithfulness, text alignment, and image realism we asked annotators to rate whether synthetic image A or B performed better, or whether they ``Tie.'' We define the metrics using the instructions seen in~\autoref{fig:metric-definitions}.

We also gave annotators the option to mark whether one of the synthetic image makes no substantial edit to the original object (i.e. the image copies and recolors the same object), which would be unfairly marked as having better shape faithfulness and image realism at the cost of text alignment under our evaluation standard. We removed these marked samples from our final comparison, resulting in the removal of less than $10\%$ of samples from the $2500$ total samples (from 25 annotators rating 100 images each) across all evaluations.

\section{\label{sec:ablations} Additional Ablations}
We report an ablation study in  \autoref{tab:ablation_val}. (1) uses a standard guidance scale of $7.5$ and copies the background of the real image onto the prediction at each timestep as done in Blended Diffusion~\cite{avrahami2022blended}, (2) uses DDIM inverted noise during the generation process, (3) re-weights cross-attention maps based on the change between $\mathcal{P}_{src}$ and $\mathcal{P}_{edit}$ as done in P2P~\cite{hertz2022prompt}\footnote{When re-weighting, we use a constant scalar upweighting of 2.5 as determined by hyperparameter sweeps in early experiments.}, (4) applies Inside-Outside Attention only to the cross-attention layers (Token Inside-Outside Attn), (5) applies Inside-Outside Attention with a hard mask for the cross-attention and soft mask for the self-attention layers (Soft Inside-Outside Attn), and (6) applies Inside-Outside Attention with a hard mask for both the cross- and self-attention layers (Hard Inside-Outside Attn), the design used in our final method. Comparing (1) and (2), we show that using DDIM inverted noise helps in both mIoU and FID. For (3), we empirically find that higher guidance scale, when used in combination with DDIM inversion, makes the model rely less on the the inverted noise, resulting in less realistic editing. However, we find that simply lowering the guidance scale leads to degradation in text faithfulness. Using cross-attention re-weighting mitigates this issue and allows us to achieve better image realism with similar performance in shape and text faithfulness. In (4), when we apply the Inside-Outside Attention mechanism only to the cross-attention layers we observe a small boost in mIoU with the same FID score, and when we apply it to both the cross- and self- attention layers in (5) we observe a more significant boost of 6.6 points in mIoU and 0.4 points in FID from (3). Comparing (5) and (6) we find that using a hard mask for Inside-Outside Attention performs better than using a soft mask, as seen by the further boost of 1.4 points in mIoU and 1.2 points in FID. Our final method that combines DDIM inversion, re-weighting, and hard Inside-Outside Attention achieves the best performance in mIoU and FID with scores of 67.6 and 39.0 respectively without a degradation in CLIP score. In~\autoref{fig:copy-background} we also demonstrate that our Inside-Outside Attention Mechanism is able to provide stronger shape signal than ``copy background.'' Specifically, ``copy background'' provides a weaker shape cue because its signal is centered around how well the edit blends with the copied background at each timestep, which is harder to determine at early and noisy timesteps.

\section{\label{sec:success_failure} Success / Failure Cases}
\begin{figure*}
\begin{scriptsize}
\begin{center}
\includegraphics[width=0.70\linewidth]{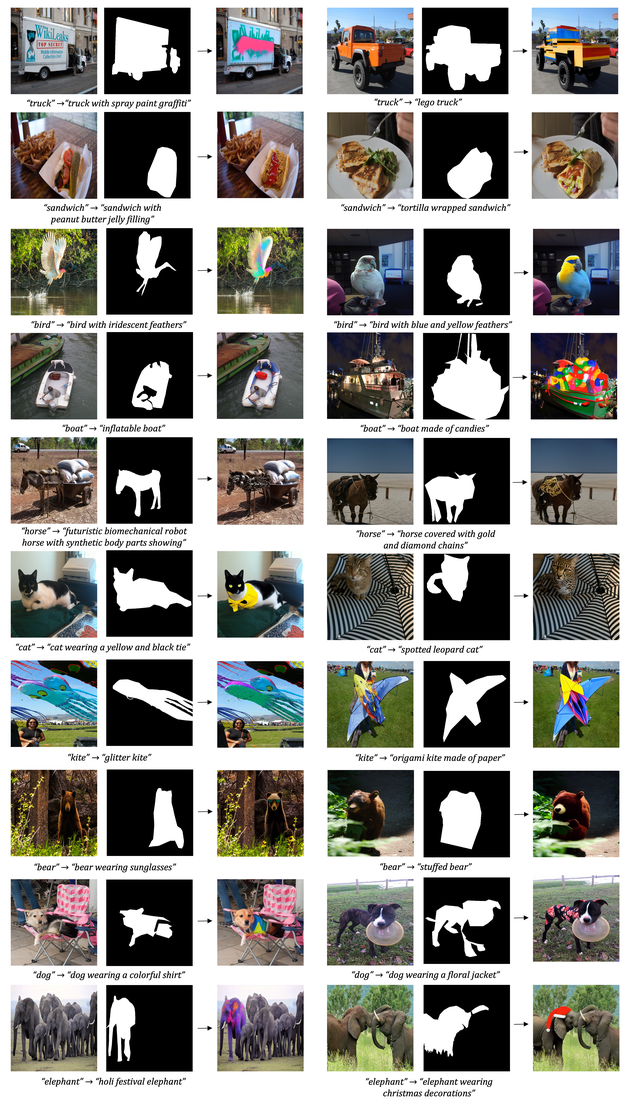}
\end{center}
\caption{Examples of success cases from our method that demonstrate its ability to handle partially occluded masks, add accessories, transform materials, or recolor objects.}
\label{fig:successes}
\end{scriptsize}
\end{figure*}

\begin{figure*}
\begin{scriptsize}
\begin{center}
\includegraphics[width=0.85\linewidth]{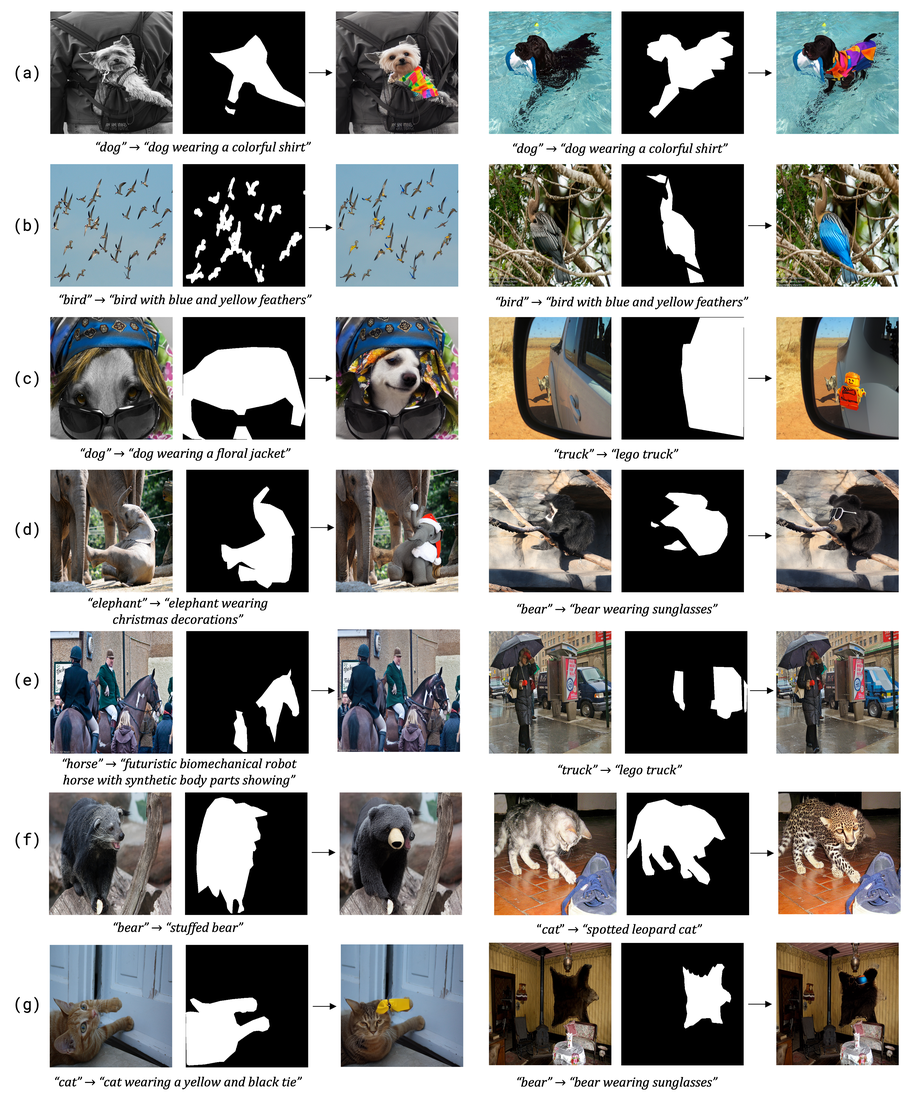}
\end{center}
\caption{Examples of failure cases from our method that relate to (a) uncommon pose, (b) uncommon perspective, (c) multi-part mask, (d) ghosting, (e) global context, (f) accessory placement.}
\label{fig:failures}
\end{scriptsize}
\end{figure*}

\noindent\textbf{Success Cases} In \autoref{fig:successes} we show edits made by our method for each prompt in the MS-COCO ShapePrompts benchmark. We demonstrate that our method is able to handle partially occluded masks and maintain relationships between the object and background, as seen in the case of the ``inflatable boat'' where the edit maintains the position of the man and dog on that boat. We demonstrate that our method is able to add accessories while simultaneously respecting the input shape, as seen by the ``elephant wearing christmas decorations'' where an ear is converted to a santa hat. Finally, our method is seamlessly able to edit material (``lego truck'', ``boat made of candies'', ``origami kite made of paper'') and color (``truck with spray paint graffiti'', ``bird with iridescent feathers'', ``holi festival elephant''). \\
\noindent\textbf{Failure Cases} We also show failure cases of our method in \autoref{fig:failures}. Sometimes the shape is inherently difficult, such as the case with (a) uncommon pose (i.e., our method repositions an elephant sitting on its hind legs to standing), (b) uncommon perspective (i.e., our method transforms a close-up of a dog's eyes to a dog's entire face and converts its hair into fabric to obey the ``floral jacket'' in the prompt), or (c) multi-part mask inputs (i.e., our method only edits the front half of the truck into a lego material). Since we use DDIM inversion (d) ghosting can occur where remnants of the original object (e.g. a mouth or ear) can appear in the synthesized object. Since we localize the attention maps (e) global context can be ignored (i.e., our method edits a colorful dog into a black-and-white photo or creates artifacts at the boundary between a dog's legs and water). Finally, our method may produce strange (f) accessory placement (i.e., our method places a bowtie on the cat's arm because its neck is not visible in the original image).

\section{\label{sec:additional_settings} Additional Editing Results}
\begin{figure*}[h!]
    \renewcommand{\arraystretch}{0.3}
    \setlength\tabcolsep{1.5pt}
    \resizebox{\linewidth}{!}{
    \begin{tabular}{cccccc}
         \textbf{Image} & 
         \textbf{Inferred Shape} & 
         \textbf{Ours} & 
         \textbf{Image} & 
         \textbf{Inferred Shape} & 
         \textbf{Ours} 
         \\ \\
        \includegraphics[width=0.17\textwidth]{figures/automatic_mask_samples/179298_real.png} &
        \includegraphics[width=0.17\textwidth]{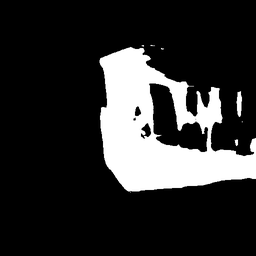} &
        \includegraphics[width=0.17\textwidth]{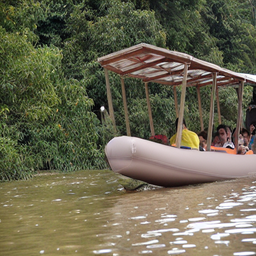} &
        \includegraphics[width=0.17\textwidth]{figures/automatic_mask_samples/3218_real.png} &
        \includegraphics[width=0.17\textwidth]{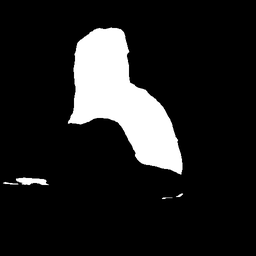} &
        \includegraphics[width=0.17\textwidth]{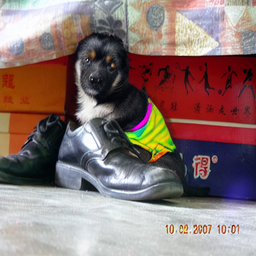} \\
        \multicolumn{3}{c}{\emph{``boat'' $\rightarrow$ ``inflatable boat''}} &
        \multicolumn{3}{c}{\emph{``dog''$\rightarrow$ ``dog wearing a colorful shirt''}} \\ \\
        \includegraphics[width=0.17\textwidth]{figures/automatic_mask_samples/588006_real.png} &
        \includegraphics[width=0.17\textwidth]{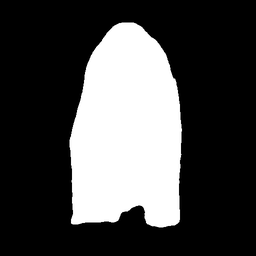} &
        \includegraphics[width=0.17\textwidth]{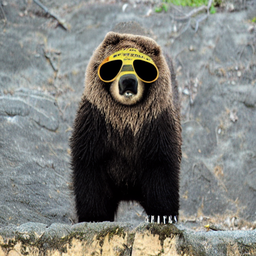} &
        \includegraphics[width=0.17\textwidth]{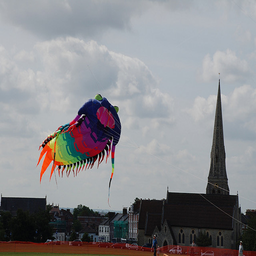} &
        \includegraphics[width=0.17\textwidth]{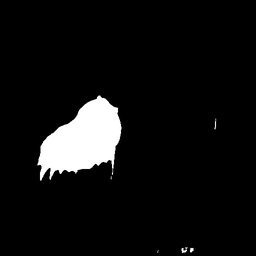} &
        \includegraphics[width=0.17\textwidth]{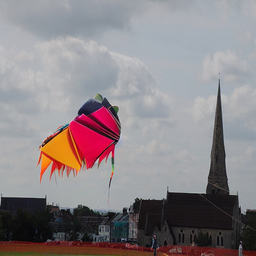} \\
        \multicolumn{3}{c}{\emph{``bear'' $\rightarrow$ ``bear wearing sunglasses''}} &
        \multicolumn{3}{c}{\emph{``kite''$\rightarrow$ ``origami kite made of paper''}} \\ \\
        \includegraphics[width=0.17\textwidth]{figures/automatic_mask_samples/1197331_real.png} &
        \includegraphics[width=0.17\textwidth]{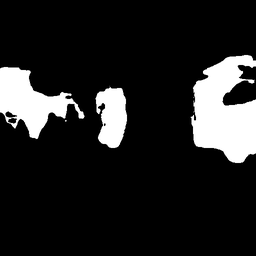} &
        \includegraphics[width=0.17\textwidth]{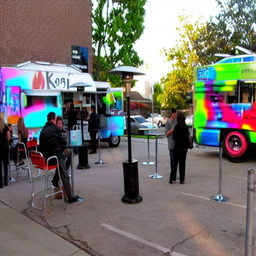} &
        \includegraphics[width=0.17\textwidth]{figures/automatic_mask_samples/586917_real.png} &
        \includegraphics[width=0.17\textwidth]{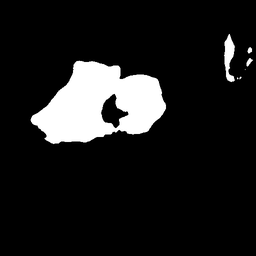} &
        \includegraphics[width=0.17\textwidth]{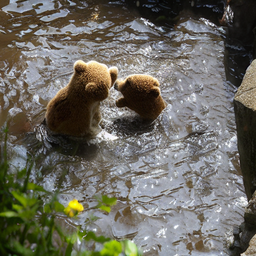} \\
        \multicolumn{3}{c}{\emph{``truck'' $\rightarrow$ ``truck with spray paint graffiti''}} &
        \multicolumn{3}{c}{\emph{``bear''$\rightarrow$ ``stuffed bear''}} \\ \\
        \includegraphics[width=0.17\textwidth]{figures/automatic_mask_samples/54758_real.png} &
        \includegraphics[width=0.17\textwidth]{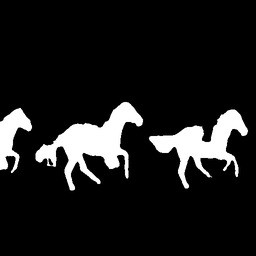} &
        \includegraphics[width=0.17\textwidth]{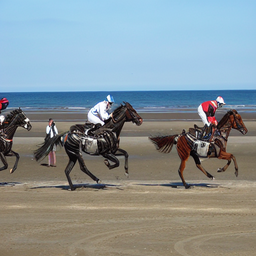} &
        \includegraphics[width=0.17\textwidth]{figures/automatic_mask_samples/12414_real.png} &
        \includegraphics[width=0.17\textwidth]{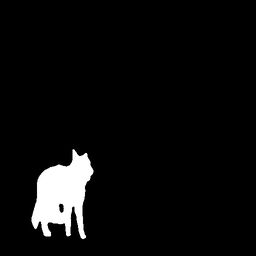} &
        \includegraphics[width=0.17\textwidth]{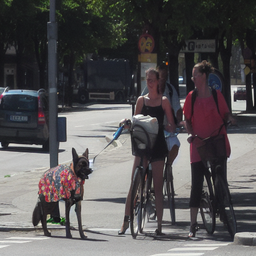} 
        \\
        \multicolumn{3}{c}{\makecell{\emph{``horse'' $\rightarrow$ ``futuristic biomechanical robot} \\ \emph{horse with synthetic body parts showing''}}} & \multicolumn{3}{c}{\emph{``dog''$\rightarrow$ ``dog wearing a floral jacket''}} \\\\
        \includegraphics[width=0.17\textwidth]{figures/automatic_mask_samples/39009_real.png} &
        \includegraphics[width=0.17\textwidth]{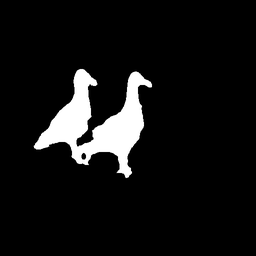} &
        \includegraphics[width=0.17\textwidth]{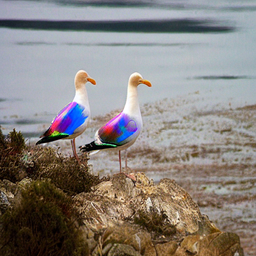} &
        \includegraphics[width=0.17\textwidth]{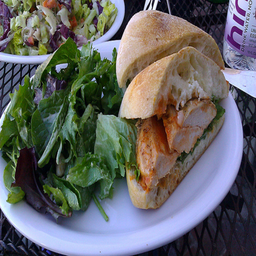} &
        \includegraphics[width=0.17\textwidth]{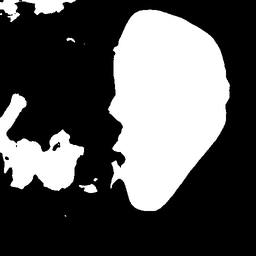} &
        \includegraphics[width=0.17\textwidth]{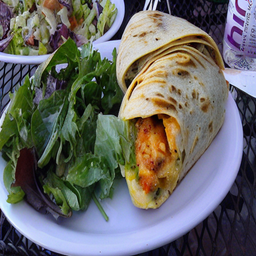}
        \\
        \multicolumn{3}{c}{\makecell{\emph{``bird''$\rightarrow$ ``bird with iridescent feathers''} \\ \emph{horse with synthetic body parts showing''}}} & \multicolumn{3}{c}{\emph{`sandwich''$\rightarrow$ ``tortilla wrapped sandwich''}} \\\\
    \end{tabular}}
    \vspace{0.3cm}
    \caption{Additional examples generated by our method with masks automatically inferred from the text (predicted by MaskFormer \cite{cheng2021maskformer}). Depending on the inferred shape, our method is able to edit multiple instances and handle complex shapes caused by noisy mask predictions or severe occlusions.}
    \label{fig:inferred_mask}
\end{figure*}

\begin{figure*}
\begin{scriptsize}
\begin{center}
\includegraphics[width=0.5\linewidth]{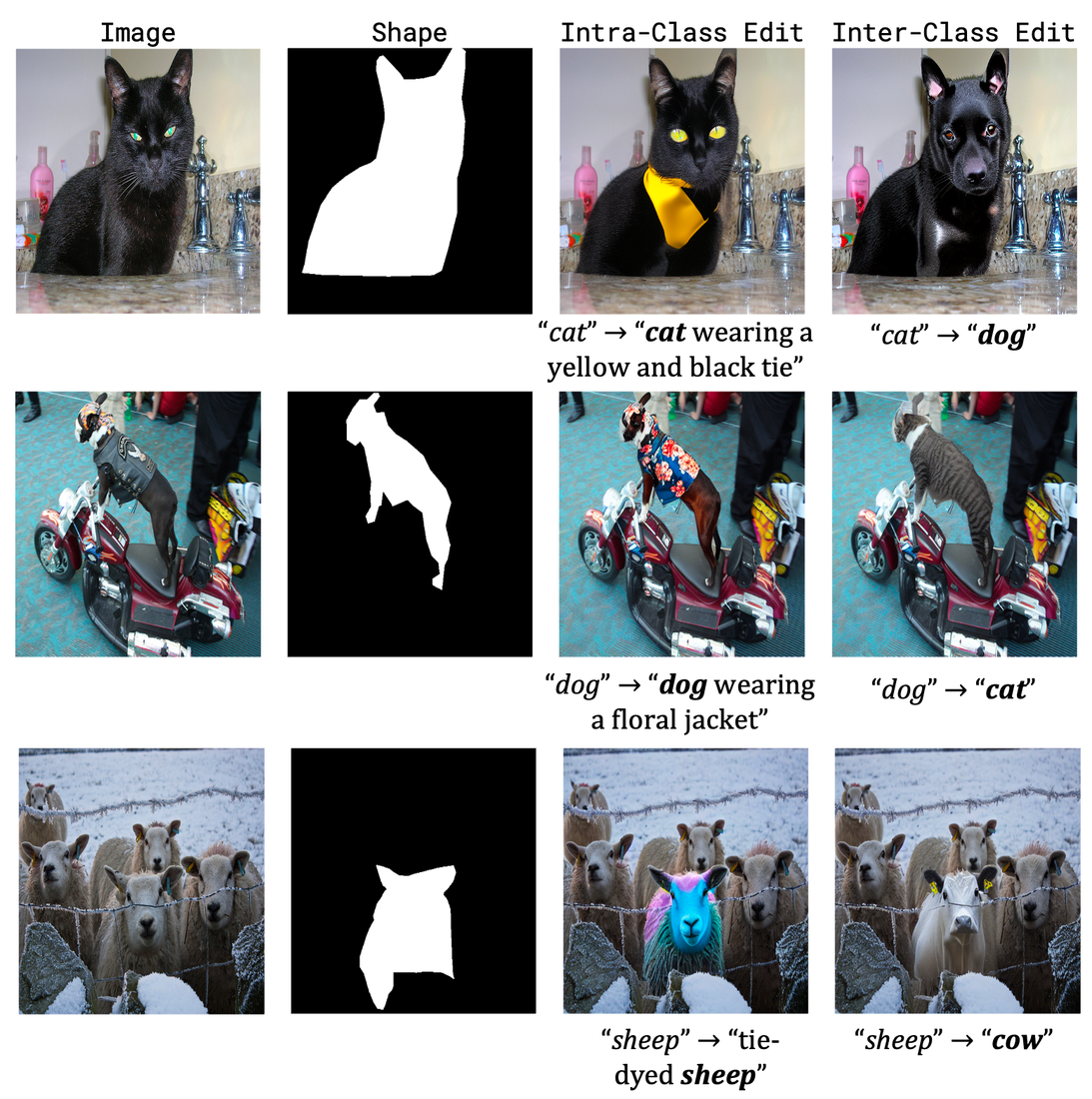}
\end{center}
\caption{Additional examples of inter-class edits.}
\label{fig:additional_inter_class}
\end{scriptsize}
\end{figure*}

\begin{figure*}
\begin{scriptsize}
\begin{center}
\includegraphics[width=\linewidth]{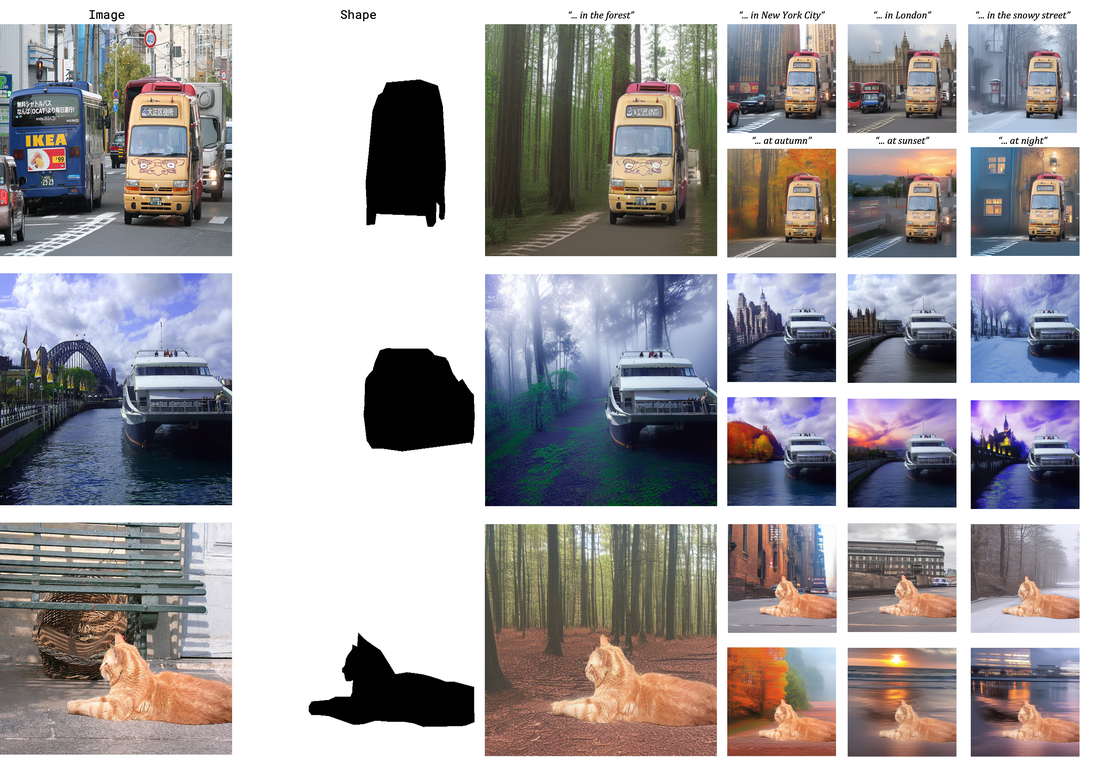}
\end{center}
\caption{Additional examples of outside edits from our method where we transform the background to various locations (New York City, London), seasons (winter, autumn), and times of day (sunset, night) for various objects (truck, boat, cat).}
\label{fig:background-edit-supp}
\end{scriptsize}
\end{figure*}

\begin{figure*}
\begin{scriptsize}
\begin{center}
\includegraphics[width=\linewidth]{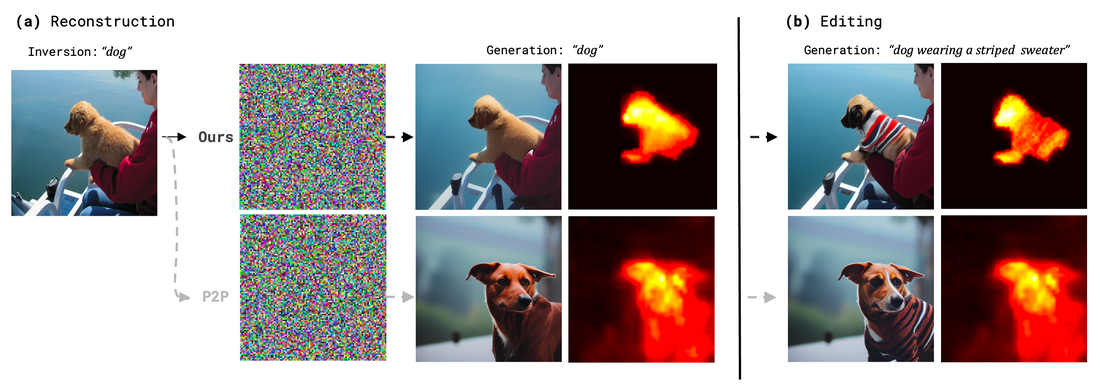}
\end{center}
\caption{Spurious attentions and classifier-free guidance also affects P2P \cite{hertz2022prompt}. We compare our method (top) and P2P (bottom) for reconstructing (left) and editing (right) an image with corresponding cross attention maps for the token ``dog'' averaged over all layers and timesteps.}
\label{fig:p2p-intuition}
\end{scriptsize}
\end{figure*}

\noindent\textbf{Inferred Shape Edits} We show additional examples of edits made by our method with an inferred shape as input in~\autoref{fig:inferred_mask}. Our method is able to handle a wide array of inferred shapes including those with multiple instances, noise, and occlusions.
\\
\noindent\textbf{Inter-Class Edits} We show additional examples of inter-class edits in~\autoref{fig:additional_inter_class}, including converting from cat to dog, dog to cat, or sheep to cow. \\\noindent\textbf{Outside Edits.} We show additional examples of outside edits in~\autoref{fig:background-edit-supp}, including transforming the background to different locations, seasons, or times of day. \\
\noindent\textbf{Spurious Attentions and Classifier-Free Guidance}
In the main text we discuss how our Inside-Outside Attention mechanism is able to better perform reconstruction and editing with classifier-free guidance by removing spurious attentions. We additionally show our method vs. P2P~\cite{hertz2022prompt} in the same setting in~\autoref{fig:p2p-intuition}. P2P exhibits spurious attentions where the token ``dog'' not only attends to the dog but also the background, causing the shape of the original dog to diverge completely.

\begin{figure*}[h!]
    \renewcommand{\arraystretch}{0.3}
    \setlength\tabcolsep{1.5pt}
    \resizebox{\linewidth}{!}{
    \begin{tabular}{cccccc}
         \textbf{Image} & 
         \textbf{P2P} & 
         \textbf{NTI + P2P} & 
         \textbf{Plug-And-Play} &
         \textbf{InstructPix2Pix} &
         \textbf{Ours}
         \\ \\
         \includegraphics[width=0.17\textwidth]{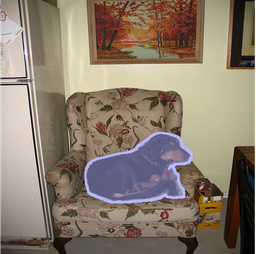} &
        \includegraphics[width=0.17\textwidth]{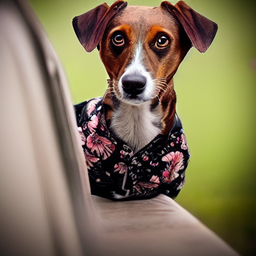} &
        \includegraphics[width=0.17\textwidth]{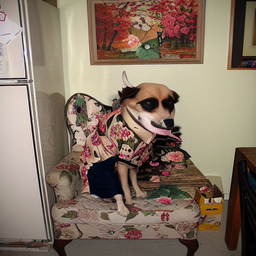} &
        \includegraphics[width=0.17\textwidth]{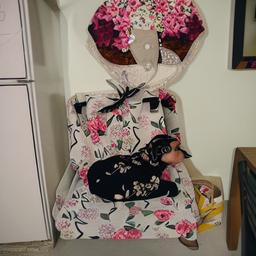} &
        \includegraphics[width=0.17\textwidth]{figures/concurrent_samples/8627_instructpix2pix.png} &
        \includegraphics[width=0.17\textwidth]{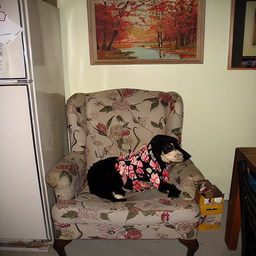} \\
        \multicolumn{6}{c}{\emph{``dog'' $\rightarrow$ ``dog wearing a floral jacket''}} \\ \\
        \includegraphics[width=0.17\textwidth]{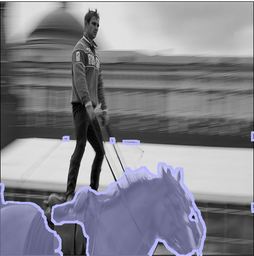} &
        \includegraphics[width=0.17\textwidth]{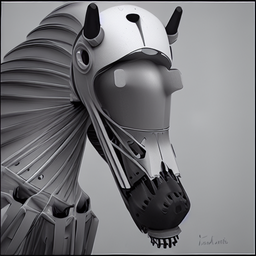} &
        \includegraphics[width=0.17\textwidth]{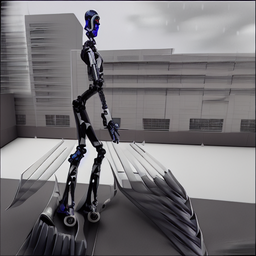} &
        \includegraphics[width=0.17\textwidth]{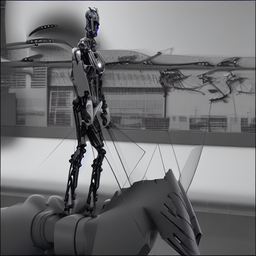} &
        \includegraphics[width=0.17\textwidth]{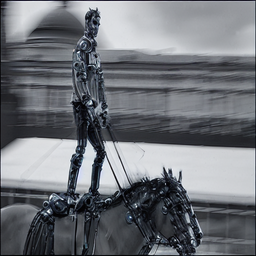} &
        \includegraphics[width=0.17\textwidth]{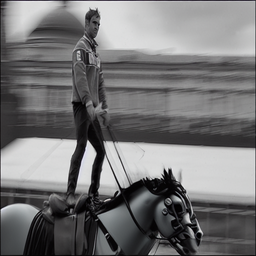} \\
        \multicolumn{6}{c}{\emph{``horse'' $\rightarrow$``futuristic biomechanical robotic horse with synthetic body parts showing''}} \\ \\
        \includegraphics[width=0.17\textwidth]{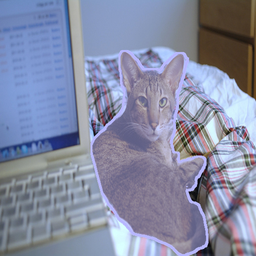} &
        \includegraphics[width=0.17\textwidth]{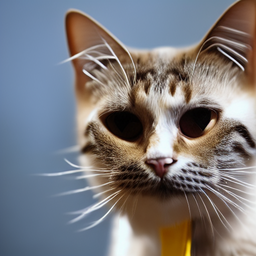} &
        \includegraphics[width=0.17\textwidth]{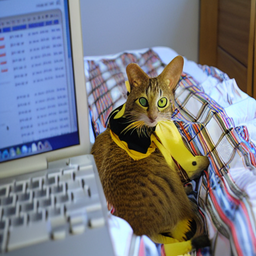} &
        \includegraphics[width=0.17\textwidth]{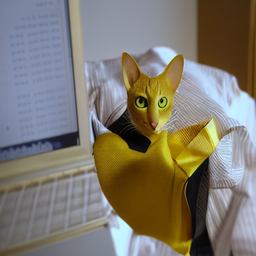} &
        \includegraphics[width=0.17\textwidth]{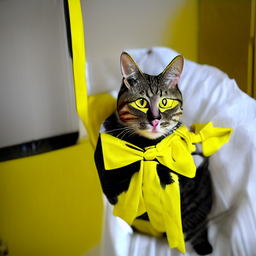} &
        \includegraphics[width=0.17\textwidth]{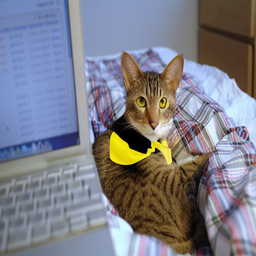} \\
        \multicolumn{6}{c}{\emph{``cat'' $\rightarrow$ ``cat wearing a yellow and black tie''}} \\ \\
        \includegraphics[width=0.17\textwidth]{figures/concurrent_samples/7385_real.png} &
        \includegraphics[width=0.17\textwidth]{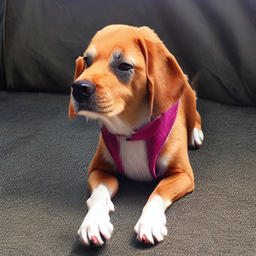} &
        \includegraphics[width=0.17\textwidth]{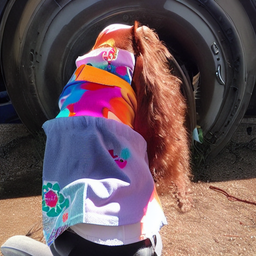} &
        \includegraphics[width=0.17\textwidth]{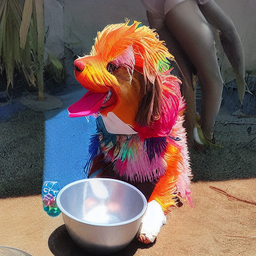} &
        \includegraphics[width=0.17\textwidth]{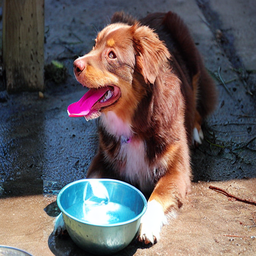} &
        \includegraphics[width=0.17\textwidth]{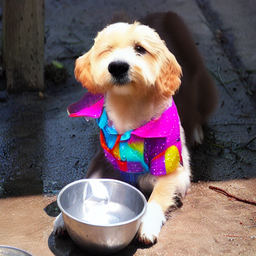} \\
        \multicolumn{6}{c}{\emph{``dog wearing a colorful shirt''}} \\ \\
        \includegraphics[width=0.17\textwidth]{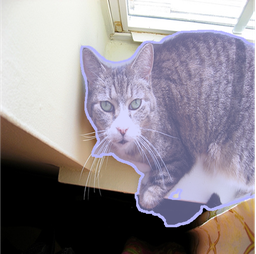} &
        \includegraphics[width=0.17\textwidth]{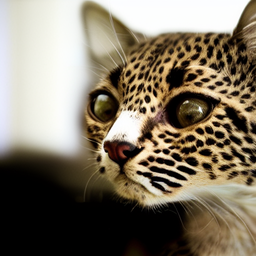} &
        \includegraphics[width=0.17\textwidth]{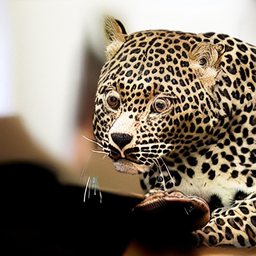} &
        \includegraphics[width=0.17\textwidth]{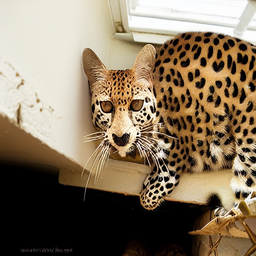} &
        \includegraphics[width=0.17\textwidth]{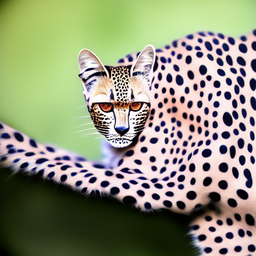} &
        \includegraphics[width=0.17\textwidth]{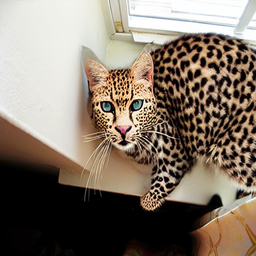} \\
        \multicolumn{6}{c}{\emph{``cat'' $\rightarrow$ ``spotted leopard cat''}} \\ \\
    \end{tabular}}
    \vspace{0.2cm}
    \caption{Qualitative examples comparing our method with concurrent work for structure preserving editing. We compare against P2P~\cite{hertz2022prompt}, NTI + P2P~\cite{mokady2022nulltext}, Plug-and-Play~\cite{tumanyan2022pnp}, and InstructPix2Pix~\cite{brooks2022instructpix2pix}. Here, we use the variant of our method that uses inferred shape, which requires the same amount of input (real image and text prompts) as the structure preserving methods.}
    \label{fig:concurrent}
\end{figure*}

\end{document}